%% file: main.tex
\documentclass{article}

\PassOptionsToPackage{compress}{natbib}
\usepackage[preprint]{corl_2022} 

\usepackage{url}
\usepackage{setspace}
\usepackage{graphicx}
\usepackage{xspace}
\usepackage{enumitem}
\usepackage{wrapfig}
\usepackage{booktabs} 
\usepackage{multirow}
\usepackage{floatrow}
\usepackage{algorithm}
\usepackage{algorithmic}
\usepackage{amsmath,amsfonts,bm}
\usepackage{lineno}
\usepackage{amssymb}
\usepackage{caption,subcaption, overpic, textpos}
\usepackage{lscape}

\usepackage{amssymb}
\usepackage{pifont}
\usepackage{fontawesome}
\usepackage{adjustbox}
\usepackage{array}
\newcolumntype{R}[2]{%
    >{\adjustbox{angle=#1,lap=\width-(#2)}\bgroup}%
    l%
    <{\egroup}%
}

\newcommand{\figvspace}{-4mm}
\usepackage{rotating}
\definecolor{forestgreen}{rgb}{0.13, 0.55, 0.13}
\definecolor{indiagreen}{rgb}{0.07, 0.53, 0.03}

\definecolor{MyDarkBlue}{rgb}{0,0.08,1}
\definecolor{MyDarkGreen}{rgb}{0.02,0.6,0.02}
\definecolor{MyDarkRed}{rgb}{0.8,0.02,0.02}
\definecolor{MyDarkOrange}{rgb}{0.40,0.2,0.02}
\definecolor{MyPurple}{RGB}{111,0,255}
\definecolor{MyRed}{rgb}{1.0,0.0,0.0}
\definecolor{MyGold}{rgb}{0.75,0.6,0.12}
\definecolor{MyDarkgray}{rgb}{0.66, 0.66, 0.66}

\newlength\savewidth
\newcommand{\tablestyle}[2]{\setlength{\tabcolsep}{#1}\renewcommand{\arraystretch}{#2}\centering\footnotesize}
\definecolor{baselinecolor}{HTML}{d6eaf8}

\newcommand{\mgpt}{MaskViT\xspace}
\newcommand{\sw}{$1\times16\times16$}
\newcommand{\stw}[1]{$16\times#1\times#1$}
\newcommand{\website}{https://maskedvit.github.io/}

\title{\mgpt: Masked Visual Pre-Training\\for Video Prediction}

\author{
  Agrim Gupta$^1$
  \And
  Stephen Tian$^1$
  \And
  Yunzhi Zhang$^1$
  \And
  Jiajun Wu$^1$
  \AND
  {Roberto Mart\'in-Mart\'in$^{2,1}$
  ~~~~~~~~~
  Li Fei-Fei$^1$} \\ \\
  {$^1$ Stanford University, $^2$ Salesforce AI}
}
\begin{document}
\maketitle


\begin{abstract}
The ability to predict future visual observations conditioned on past observations and motor commands can enable embodied agents to plan solutions to a variety of tasks in complex environments. This work shows that we can create good video prediction models by pre-training transformers via masked visual modeling. Our approach, named \mgpt, is based on two simple design decisions. First, for memory and training efficiency, we use two types of window attention: spatial and spatiotemporal. Second, during training, we mask a \textit{variable} percentage of tokens instead of a \textit{fixed} mask ratio. For inference, \mgpt generates all tokens via iterative refinement where we incrementally decrease the masking ratio following a mask scheduling function. On several datasets we demonstrate that \mgpt outperforms prior works in video prediction, is parameter efficient, and can generate high-resolution videos ($256\times256$). Further, we demonstrate the benefits of inference speedup (up to $512\times$) due to iterative decoding by using \mgpt for planning on a real robot. Our work suggests that we can endow embodied agents with powerful predictive models by leveraging the general framework of masked visual modeling with minimal domain knowledge. 
\end{abstract}



\input{01-intro}

\input{02-rw}
\input{03-method}
\input{04-exp}

\section{Conclusion}
\label{sec:conclusion}
In this work, we explore \mgpt, a simple method for video prediction which leverages masked visual modeling as a pre-training task and transformers with window attention as a backbone for computation efficiency. We showed that by masking a \textit{variable} number of tokens during training, we can achieve competitive video prediction results. Our iterative decoding scheme is significantly faster than autoregressive decoding and enables planning for real robot manipulation tasks. 

\textbf{Limitations and future work.} While our results are encouraging, we found that using per frame quantization can lead to flicker artifacts, especially in videos that have a static background like RoboNet. Although \mgpt is efficient in terms of memory and parameters, scaling up video prediction, especially for scenarios that have significant camera motion (e.g., self-driving \citep{geiger2013vision} and egocentric videos \citep{grauman2021ego4d}) remains challenging. Finally, an important future avenue of exploration is scaling up the complexity of robotic tasks~\citep{srivastava2022behavior} integrating our video prediction method in more complex planning algorithms.


\clearpage
\acknowledgments{We thank Abhishek Kadian for helpful discussions. This work was supported by Department of Navy award (N00014-19-1-2477) issued by the Office of Naval Research and Stanford HAI. ST is supported by the NSF Graduate Research Fellowship under Grant No. DGE-1656518.}


\bibliography{main}  

\clearpage

\input{05-appendix}

\end{document}

%% file: 01-intro.tex
\section{Introduction}

Evidence from neuroscience suggests that human cognitive and perceptual capabilities are supported by a predictive mechanism to anticipate future events and sensory signals~\cite{tanji1976anticipatory, wolpert1995internal}. Such a mental model of the world can be used to simulate, evaluate, and select among different possible actions. This process is fast and accurate, even under the computational limitations of biological brains~\cite{wu2016capacity}. Endowing robots with similar predictive capabilities would allow them to plan solutions to multiple tasks in complex and dynamic environments, e.g., via visual model-predictive control~\cite{finn2017deep,ebert2018visual,hirose2019deep}.

Predicting visual observations for embodied agents is however challenging and computationally demanding: the model needs to capture the complexity and inherent stochasticity of future events while maintaining an inference speed that supports the robot's actions. 
Therefore, recent advances in autoregressive generative models, which leverage Transformers~\cite{vaswani2017attention} for building neural architectures and learn good representations via self-supervised generative pretraining~\cite{devlin2019bert}, have not benefited video prediction or robotic applications. 
We in particular identify three technical challenges. First, memory requirements for the full attention mechanism in Transformers scale quadratically with the length of the input sequence, leading to prohibitively large costs for videos. Second, there is an inconsistency between the video prediction task and autoregressive masked visual pretraining -- while the training process assumes \textit{partial} knowledge of the ground truth future frames, at test time the model has to predict a complete sequence of future frames from \textit{scratch}, leading to poor video prediction quality. Third, the common autoregressive paradigm effective in other domains would be too slow for robotic applications.

To address these challenges, we present \textbf{Mask}ed \textbf{Vi}deo \textbf{T}ransformers (\mgpt): a simple, effective and scalable method for video prediction based on masked visual modeling. Since using pixels directly as frame tokens would require an inordinate amount of memory, we use a discrete variational autoencoder (dVAE)~\citep{van2017neural, esser2021taming} that compresses frames into a smaller grid of visual tokens. We opt for compression in the spatial (image) domain instead of the spatiotemporal domain (videos), as preserving the correspondence between each original and tokenized video frame allows for flexible conditioning on any subset of frames -- initial (past), final (goal),  and possibly equally spaced intermediate frames. However, despite operating on tokens, representing $16$ frames at $256$ tokens per frame still requires $4,096$ tokens, incurring prohibitive memory requirements for full attention. Hence, to further reduce memory, \mgpt is composed of alternating transformer layers with non-overlapping \textit{window-restricted}~\citep{vaswani2017attention} spatial and spatiotemporal attention.

\begin{figure}[t]
\centering
\includegraphics[width=\textwidth]{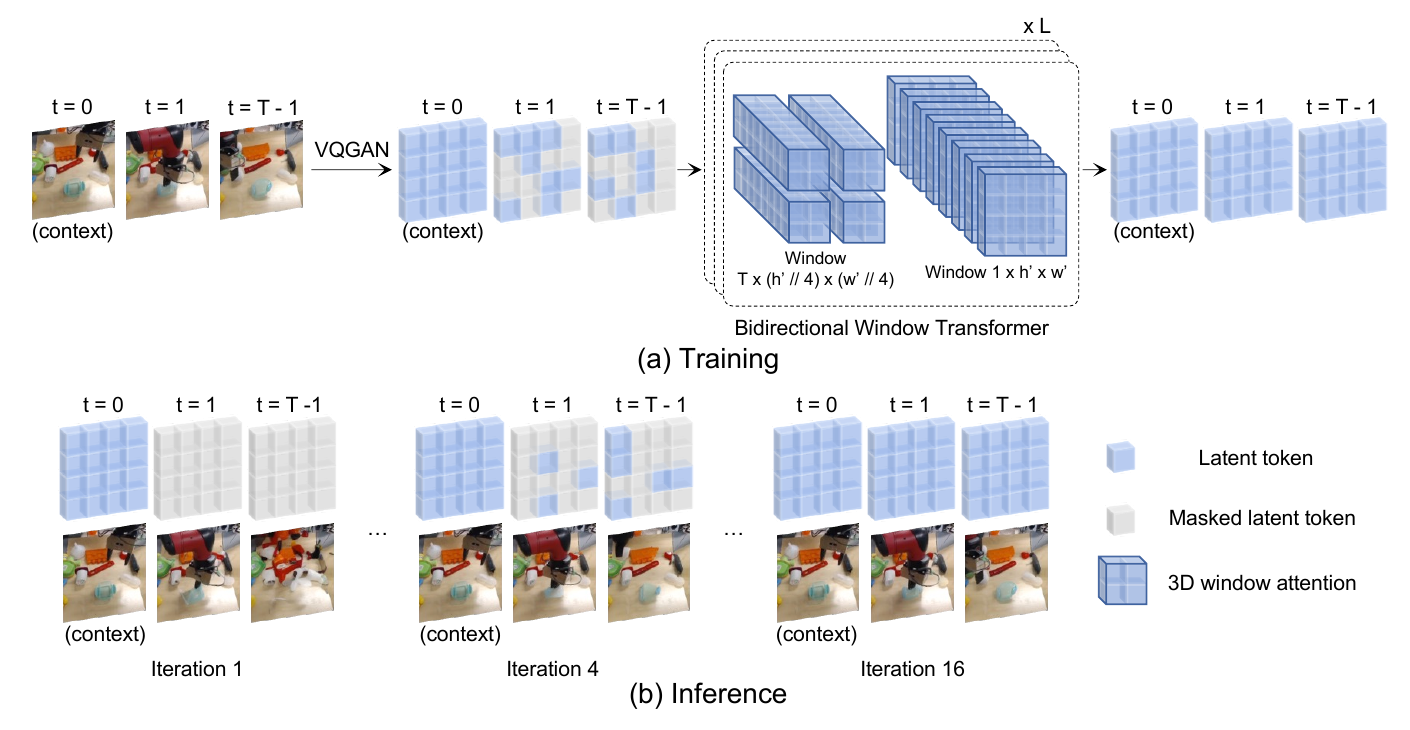}
  \caption{\textbf{\mgpt}. (a) Training: We first encode the video frames into latent codes via VQ-GAN. A \textit{variable} number of tokens in future frames are masked, and the network is trained to predict the masked tokens. A block in \mgpt consists of two layers with window-restricted attention: spatial and spatiotemporal. (b) Inference: Videos are generated via iterative refinement where we incrementally decrease the masking ratio following a mask scheduling function. Videos available at \href{\website}{this project page}.\vspace{-15pt}}
\vspace{\figvspace}
\label{fig:model}
\end{figure}

To reduce the inconsistency between the masked pretraining and the video prediction task and to speed up inference, we take inspiration from non-autoregressive, iterative decoding methods in generative algorithms from other domains~\citep{sohl2015deep, ho2020denoising, nichol2021improved, ghazvininejad2019mask, chang2022maskgit}. We propose a novel iterative decoding scheme for videos based on a mask scheduling function that specifies, during inference, the number of tokens to be decoded and kept at each iteration. A few initial tokens are predicted over multiple initial iterations, and then the majority of the remaining tokens can be predicted rapidly over the final few iterations. This brings us closer to the ultimate video prediction task, where only the first frame is known and all tokens for other frames must be inferred. 
Our proposed prediction procedure provides fast predictions without temporally increasing quality degradation due to its iterative non-autoregressive nature. 
To further close the training-test gap, during training we mask a \textit{variable} percentage of tokens, instead of using a \textit{fixed} masking ratio. This simulates the different masking ratios \mgpt will encounter during iterative decoding in the actual video prediction task.

Through experiments on several publicly available real-world video prediction datasets~\cite{ebert2017self, geiger2013vision, dasari2019robonet}, we demonstrate that \mgpt achieves competitive or state-of-the-art results in a variety of metrics. Moreover, \mgpt can predict considerably higher resolution videos ($256 \times 256$) than previous methods. More importantly, thanks to iterative decoding, \mgpt is up to $512 \times$ faster than autoregressive methods, enabling its application for planning on a real robot (\S~\ref{subsec:robot}). These results indicate that we can endow embodied agents with powerful predictive models by leveraging the advances in self-supervised learning in language and vision, without engineering domain-specific solutions.

%% file: 02-rw.tex
\section{Related Work}
\label{sec:related}
\textbf{Video prediction.} The video prediction task refers to the problem of generating videos conditioned on past frames~\cite{ranzato2014video, lotter2016deep}, possibly with an additional natural language description~\cite{li2018video, gupta2018imagine, pan2017create, wu2021n} and/or motor commands~\citep{finn2016unsupervised, villegas2019high, wu2021greedy, babaeizadeh2021fitvid}. Multiple classes of generative models have been utilized to tackle this problem, such as Generative adversarial networks (GANs)~\cite{clark2019adversarial,tulyakov2018mocogan, luc2020transformation}, Variational Autoencoders (VAEs)~\cite{villegas2019high, wu2021greedy, babaeizadeh2021fitvid, babaeizadeh2018stochastic,  denton2018stochastic, akan2021slamp, akan2022stochastic}, invertible networks~\citep{dorkenwald2021stochastic}, autoregressive~\cite{yan2021videogpt, rakhimov2020latent, nash2022transframer} and diffusion~\cite{ho2022video, voleti2022MCVD} models. Our work focuses on predicting future frames conditioned on past frames or motor commands and belongs to the family of two-stage methods that first encode the videos into a downsampled latent space and then use transformers to model an autoregressive prior~\cite{yan2021videogpt, rakhimov2020latent}. A common drawback of these methods is the large inference time due to autoregressive generation. \mgpt overcomes this issue by using an iterative decoding scheme, which significantly reduces inference time. 

\textbf{Masked autoencoders.} Masked autoencoders are a type of denoising autoencoder~\citep{vincent2008extracting} that learn representations by (re)generating the original input from corrupted (i.e., masked) inputs. Masked language modeling was first proposed in BERT~\citep{devlin2019bert} and has revolutionized the field of natural language processing, especially when scaled to large datasets and model sizes~\citep{brown2020language, radford2019language}. The success in NLP has also been replicated in vision by masking patches of pixels~\cite{he2021masked, dosovitskiy2020image} or masking tokens generated by a pretrained dVAE~\citep{bao2022beit, chen2020generative}. Recently, these works have also been extended to video domains to learn good representations for action recognition ~\citep{tong2022videomae, feichtenhofer2022masked}. Unlike them, we apply masked visual modeling for video prediction, and we use a \textit{variable} masking ratio during training to reduce the difference between masked pretraining and video prediction. Another related line of work is leveraging good visual representations learnt via self supervised learning methods~\citep{laskin2020curl, nair2022r3m, parisi2022unsurprising} including masked autoencoders~\cite{xiao2022masked} for motor control.

%% file: 03-method.tex
\section{\mgpt: Masked Video Transformer}
\label{sec:method}

\mgpt is the result of a two-stage training procedure~\citep{van2017neural, razavi2019generating}: First, we learn an encoding of the visual data that discretizes images into tokens based on a discrete variational autoencoder (dVAE). Next, we deviate from the common autoregressive training objective and pre-train a bidirectional transformer with window-restricted attention via \textit{masked visual modeling} (MVM).
In the following section, we describe our image tokenizer, bidirectional transformer, masked visual pre-training, and iterative decoding procedure.

\subsection{Learning Visual Tokens}
Videos contain too many pixels to be used directly as tokens in a transformer architecture. Hence, to reduce dimensionality, we first train a VQ-VAE~\citep{van2017neural} for individual video frames so that we can represent videos as sequences of grids of discrete tokens. VQ-VAE consists of an encoder $E(x)$ that encodes an input image $x \in \mathbb{R}^{H \times W \times 3}$ into a series of latent vectors. The vectors are discretized through a nearest neighbour look up in a codebook of quantized embeddings, $\mathcal{Z} = \{z_k\}_{k=1}^K \subset \mathbb{R}^{n_z}$. A decoder $D$ is trained to predict a reconstruction of the image, $\hat{x}$, from the quantized encodings. In our work, we leverage VQ-GAN~\citep{esser2021taming}, which improves upon VQ-VAE by adding adversarial~\citep{goodfellow2014generative} and perceptual losses~\citep{johnson2016perceptual, zhang2018unreasonable}. Each video frame is individually tokenized into a $16 \times 16$ grid of tokens, regardless of their original resolution (Fig.~\ref{fig:model}, a, left). Instead of using 3D extensions of VQ-VAE which perform spatiotemporal compression of videos~\citep{yan2021videogpt}, our per-frame compression enables us to condition on arbitrary context frames: initial, final, and possibly intermediate ones. 

\subsection{Masked Visual Modeling (MVM)}
Inspired by the success of masked language~\citep{devlin2019bert} and image~\citep{bao2022beit, he2021masked} modeling, and in the spirit of unifying methodologies across domains, we pre-train \mgpt via MVM for video prediction. Our pre-training task and masking strategy are straightforward: we keep the latent codes corresponding to context frames intact and mask a random number of tokens corresponding to future frames. The network is trained to predict masked latent codes conditioned on the unmasked latent codes. 

Concretely, we assume access to input context frames for $T_c$ time steps, and our goal is to predict $T_p$ frames during test time. We first quantize the entire video sequence into latent codes ${Z} \in \mathbb{R}^{T \times h \times w}$. Let ${Z_p} = [z_i]_{i=1}^N$ denote the latent tokens corresponding to future video frames, where $N = T_p \times h \times w$. Unlike prior work on MVM~\citep{bao2022beit, he2021masked} that uses a \textit{fixed} masking ratio, we propose to use a \textit{variable} masking ratio that reduces the gap between pre-training task and inference leading to better evaluation results (see \S~\ref{subsec:inference}). 
Specifically, during training, for each video in a batch, we first select a masking ratio $r \in [0.5, 1)$ and then randomly select and replace $\lfloor r \cdot N \rfloor$ tokens in ${Z_p}$ with a \texttt{[MASK]} token. 
The pre-training objective is to minimize the negative log-likelihood of the visual tokens given the masked video as input: $\mathcal{L}_{\text{MVM}} = - \mathop{\mathbb{E}} \limits_{x \in \mathcal{D}} \Big[ \sum_{\forall i \in N^M} \log p(z_i| Z^M_p, Z_c) \Big]$, where $\mathcal{D}$ is the training dataset, $N^M$ represents randomly masked positions, and $Z^M_p$ denotes the output of applying the mask to $Z_p$, and $Z_c$ are latent tokens corresponding to context frames. The MVM training objective is different from the causal autoregressive training objective as the conditional dependence is \textit{bidirectional}: \textit{all} masked tokens are predicted conditioned on \textit{all} unmasked tokens.

\subsection{Bidirectional Window Transformer}
Transformer models composed entirely of global self-attention modules incur significant compute and memory costs, especially for video tasks. To achieve more efficient modeling, we propose to compute self-attention in windows, based on two types of non-overlapping configurations: 1) Spatial Window (SW): attention is restricted to all the tokens within a subframe of size $1 \times h \times w$ (the first dimension is time); 2) Spatiotemporal Window (STW): attention is restricted within a 3D window of size $T \times h' \times w'$. We sequentially stack the two types of window configurations to gain both \textit{local} and \textit{global} interactions in a single block (Fig.~\ref{fig:model}, a, center) that we repeat $L$ times. Surprisingly, we find that a small window size of $h' = w' = 4$ is sufficient to learn a good video prediction model while significantly reducing memory requirements (Table~\ref{tab:window}). Note that our proposed block enjoys global interaction capabilities without requiring padding or cyclic-shifting like prior works~\citep{liu2021swin, liu2021video}, nor developing custom CUDA kernels for sparse attention~\citep{child2019generating} as both window configurations can be instantiated via simple tensor reshaping.

\subsection{Iterative Decoding}
\label{subsec:inference}
Decoding tokens autoregressively during inference is time-consuming, as the process scales linearly with the number of tokens, and this can be prohibitively large (e.g., $4,096$ for a video with $16$ frames and $256$ tokens per frame). Our video prediction training task allows us to predict future video frames via a novel iterative non-autoregressive decoding scheme: inspired by the forward diffusion process in diffusion models \citep{ho2020denoising, nichol2021improved} and the iterative decoding in generative models \citep{ghazvininejad2019mask,chang2022maskgit} we predict videos in $T$ steps where $T << N$, the total number of tokens to predict. 

Concretely, let $\gamma(t)$, where $t \in \{\frac{0}{T}, \frac{1}{T}, \dots, \frac{T-1}{T}\}$, be a mask scheduling function (Fig.~\ref{fig:mask_schedule}) that computes the mask ratio for tokens as a function of the decoding steps. We choose $\gamma(t)$ such that it is monotonically decreasing with respect to $t$, and it holds that $\gamma(0)\rightarrow1$ and $\gamma(1)\rightarrow0$ to ensure that our method converges. At $t = 0$, we start with $Z = [Z_c, Z_p]$ where all the tokens in $Z_p$ are \texttt{[MASK]} tokens. At each decoding iteration, we predict \textit{all} the tokens conditioned on \textit{all} the previously predicted tokens. For the next iteration, we mask out $n = \lceil \gamma(\frac{t}{T}) N \rceil$ tokens by keeping all the previously predicted tokens and the most confident token predictions in the current decoding step. We use the softmax probability as our confidence measure. 

%% file: 04-exp.tex
\section{Experimental Evaluation}
\label{sec:result}
In this section, we evaluate our method on three different datasets and compare its performance with prior state-of-the-art methods, using four different metrics. We also perform extensive ablation studies of different design choices, and showcase that the speed improvements due to iterative decoding enable real-time planning for robotic manipulation tasks. For qualitative results, see \S~\ref{appendix:qual_results} and videos on our \href{\website}{project website}.

\begin{table}[t]
\label{tab:results}
\centering
\vspace{0.2cm}
\begin{minipage}{.60\textwidth}
    \centering
    \scriptsize
    \begin{tabular}{lrrrrr}
    \textbf{RoboNet~\citep{dasari2019robonet}} & param. & FVD$\downarrow$ & PSNR$\uparrow$ & SSIM$\uparrow$ & LPIPS$\downarrow$ \\\midrule
    SVG~\citep{villegas2019high} & 298M & 123.2 & 23.9 & 87.8 & 0.060 \\
    GHVAE~\citep{wu2021greedy} & 599M & 95.2 & 24.7 & 89.1 & 0.036 \\
    FitVid~\citep{babaeizadeh2021fitvid} & 302M & \textbf{62.5} & \textbf{28.2} & \textbf{89.3} & \textbf{0.024} \\\midrule
    \mgpt (ours) & 257M & 133.5 & 23.2 & 80.5 & 0.042 \\
    \mgpt (ours, $256$) & 228M & 211.7 & 20.4 & 67.1 & 0.170 \\
    & & & & \\
    \textbf{KITTI~\citep{geiger2013vision}} & param. & FVD$\downarrow$ & PSNR$\uparrow$ & SSIM$\uparrow$ & LPIPS$\downarrow$ \\\midrule
    SVG~\citep{villegas2019high} & 298M & 1217.3 & 15.0 & 41.9 & 0.327 \\
    GHVAE~\citep{wu2021greedy} & 599M & 552.9 & 15.8 & 51.2 & 0.286 \\
    FitVid~\citep{babaeizadeh2021fitvid} & 302M & 884.5 & 17.1 & 49.1 & 0.217 \\\midrule
    \mgpt (ours) & 181M & \textbf{401.9} & \textbf{27.2} & \textbf{58.1} & \textbf{0.089} \\
    \mgpt (ours, $256$) & 228M & 446.1 & 26.2 & 40.7 & 0.270 \\
    \end{tabular}
\end{minipage}
\hfill
\begin{minipage}{.38\textwidth}
    \centering
    \scriptsize
    \begin{tabular}{lrr}
    \textbf{BAIR}~\citep{ebert2017self} & param. &FVD$\downarrow$ \\\midrule
    SV2P~\cite{babaeizadeh2018stochastic} & --- & 262.5 \\
    LVT~\cite{rakhimov2020latent} & --- & 125.8 \\
    SAVP~\cite{lee2018stochastic} & --- & 116.4 \\
    DVD-GAN-FP~\cite{clark2019adversarial} & --- & 109.8 \\
    VideoGPT~\cite{yan2021videogpt} & --- & 103.3 \\
    TrIVD-GAN-FP~\cite{luc2020transformation} & --- & 103.3 \\
    VT~\cite{Weissenborn2020Scaling} & 373M & 94.0 \\
    FitVid~\citep{babaeizadeh2021fitvid} & 302M & \textbf{93.6} \\\midrule
    \mgpt (ours) & 189M & \textbf{93.7} \\
    \mgpt (ours, goal cond.) & 255M & 76.9\\
    \mgpt (ours, act cond.) & 255M & 70.5 \\
    \end{tabular}
\end{minipage}
\caption{\textbf{Comparison with prior work.} We evaluate \mgpt on BAIR, RoboNet and KITTI datasets. Our method is competitive or outperforms prior work while being more parameter efficient.}
\end{table}

\begin{table*}[t]
\vspace{-.2em}
\centering
\captionsetup{font=small,labelfont=small}
\subfloat[
\textbf{Model size}. Increasing embedding dim improves FVD.
\label{tab:model_size}
]{
\centering
\begin{minipage}{0.28\linewidth}{\begin{center}
\tablestyle{3pt}{1.05}
\scriptsize
\begin{tabular}{ccc}
blocks & embd. dim & FVD$\downarrow$ \\
\midrule
6 & 768 & \cellcolor[HTML]{d6eaf8} 96.6 \\
6 & 1024 & \textbf{94.2} \\
8 & 768 & 99.3 \\
8 & 1024 & 99.5 \\
& & \\
\end{tabular}
\end{center}}\end{minipage}
}
\hspace{1em}
\subfloat[
\textbf{Spatiotemporal window size}. Smaller window size is faster, memory efficient, and achieves lower FVD scores.
\label{tab:window}
]{
\centering
\begin{minipage}{0.36\linewidth}{\begin{center}
\tablestyle{3pt}{1.05}
\scriptsize
\begin{tabular}{ccccc}
st window & FVD$\downarrow$ & train mem. & train time\\
\midrule
$16\times4\times4$ & \cellcolor[HTML]{d6eaf8} 96.6 & 7.0 GB & 12.5 hr\\
$16\times8\times8$ & \textbf{93.7} & 7.9 GB & 14.2 hr \\
$16\times16\times16$ & 96.6 & 11.6 GB & 27.9 hr \\\midrule
full self attn. & 98.2 & 16.4 GB & 40.3 hr\\
& & \\
\end{tabular}
\end{center}}\end{minipage}
}
\hspace{1em}
\subfloat[
\textbf{Mask ratio}. \textit{Variable} masking ratio works best.
\label{tab:mask_ratio}
]{
\centering
\begin{minipage}{0.20\linewidth}{\begin{center}
\tablestyle{3pt}{1.05}
\scriptsize
\begin{tabular}{cc}
mask ratio & FVD$\downarrow$ \\
\midrule
0.75 & 189.3 \\
0.90 & 124.1 \\
0.95 & 110.9 \\
0.98 & 214.4 \\\midrule
0.5 - 1  & \cellcolor[HTML]{d6eaf8} \textbf{96.6} \\
\end{tabular}
\end{center}}\end{minipage}
}
\vspace{-.1em}
\caption{\textbf{\mgpt ablation experiments} on BAIR. We compare FVD scores to ablate important design decisions with the default setting: $6$ blocks, $768$ embedding dimension (embd. dim), $1\times16\times16$ spatial window, $16\times4\times4$ saptiotemporal (st) window, and variable masking ratio. Default settings are marked in \colorbox{baselinecolor}{blue}.}
\label{tab:ablations} \vspace{-.5em}
\end{table*}
\subsection{Experimental Setup}
\textbf{Implementation.} Our transformer model is a stack of $L$ blocks, where each block consists of two transformer layers with attention restricted to the window size of $1\times16\times16$ (spatial window) and $T\times4\times4$ (spatiotemporal window), unless otherwise specified. We use learnable positional embeddings, which are the sum of space and time positional embeddings. See \S~\ref{appendix:hyper} for architecture details and hyperparameters.

\textbf{Metrics.} We use four evaluation metrics to compare our method with prior work: Fréchet Video Distance (FVD)~\citep{unterthiner2018towards}, Peak Signal-to-noise Ratio (PSNR), Structural Similarity Index Measure (SSIM)~\citep{wang2004image} and Learned Perceptual Image Patch Similarity (LPIPS)~\citep{zhang2018unreasonable}. 
To account for the stochastic nature of video prediction, we follow prior work~\citep{babaeizadeh2021fitvid, villegas2019high} and report the best SSIM, PSNR, and LPIPS scores over $100$ trials for each video. For FVD, we use all $100$ with a batch size of 256. We only conducted $1$ trial per video for evaluating performance on the BAIR dataset~\citep{ebert2017self}.

\subsection{Comparison with Prior Work} \label{subsec:exp_comp_prior_work}
\textbf{BAIR.} We first evaluate our model on the BAIR robot pushing dataset~\citep{ebert2017self}, one of the most studied video modeling datasets. 
We follow the evaluation protocol of prior works and predict $15$ video frames given $1$ context frame and no actions. The lack of action conditioning makes this task extremely challenging and tests the model's ability to predict plausible future robot trajectories and object interactions. \mgpt achieves similar performance to FitVid~\citep{babaeizadeh2021fitvid} while being more parameter efficient, and it outperforms all other prior works. In addition, we can easily adapt \mgpt to predict goal-conditioned futures by including the last frame in $Z_c$. We find that goal conditioning significantly improves performance (approx. 18\% FVD improvement). Finally, to predict action-conditioned future frames, we linearly project the action vectors and add them to ${Z}$. As expected, action conditioning performs the best, with almost a 25\% improvement in FVD.  

\textbf{KITTI.} The KITTI dataset~\citep{geiger2013vision} is a relatively small dataset of $57$ training videos.
We follow the evaluation protocol of prior work~\citep{villegas2019high} and predict $25$ video frames given $5$ context frames. Compared to other datasets in our evaluation, KITTI is especially challenging, as it involves dynamic backgrounds, limited training data, and long-horizon predictions. We use color jitter and random cropping data augmentation for training VQ-GAN and do not use any data augmentation for training the second stage. Across all metrics, we find that \mgpt is significantly better than prior works while using fewer parameters. Training a transformer model with full self-attention would require prohibitively large GPU memory due to the long prediction horizon ($30 \times 16 \times 16 = 7680$). However, \mgpt can attend to all tokens because its spatiotemporal windows significantly reduce the size of the attention context. We also report video prediction results for the KITTI dataset at $256\times256$ resolution, a higher resolution that prior work was not able to obtain.

\textbf{RoboNet.} RoboNet~\citep{dasari2019robonet} is a large dataset of $15$ million video frames of $7$ different robotic arms interacting with objects and provides 5 dimensional robot action annotations. We follow the evaluation protocol of prior work~\citep{babaeizadeh2021fitvid} and predict $10$ video frames given $2$ context frames and future actions. At $64\times64$ resolution, \mgpt is competitive but does not outperform prior works. FVD of the VQ-GAN reconstructions is a lower bound for \mgpt. We found flicker artifacts in the VQ-GAN reconstructions, probably due to our use of per-frame latents, resulting in a high FVD score of $121$ for the VQ-GAN reconstructions. \mgpt achieves FVD scores very close to this lower bound but performs worse than prior works due to temporally inconsistent VQ-GAN reconstructions.  Finally, we also report video prediction results for the RoboNet dataset at $256\times256$ resolution.

\begin{figure*}[t]
\centering
\includegraphics[width=\linewidth]{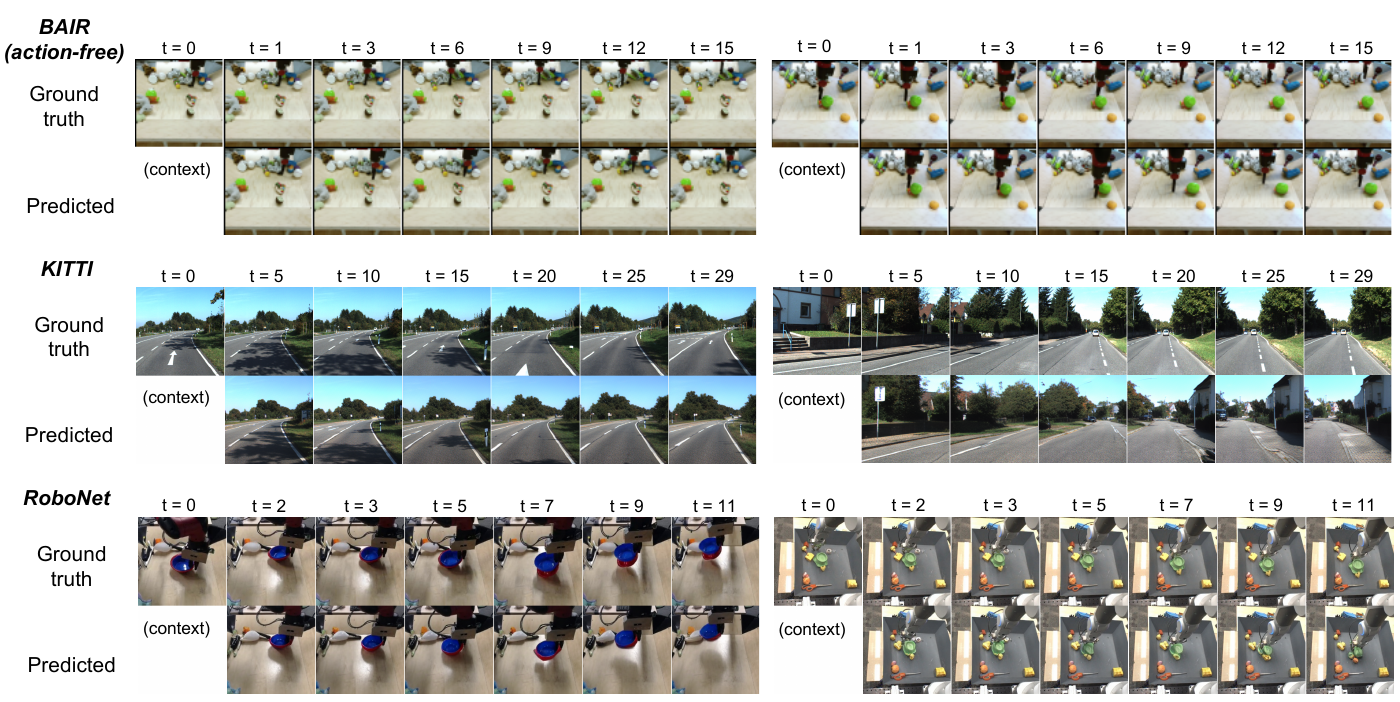}
  \caption{\textbf{Qualitative evaluation.} Video prediction results on test set of BAIR ($64\times64$), KITTI ($256\times256$), and RoboNet ($256\times256$). Zoom in for details.}
\label{fig:qual_kitti}
\end{figure*}

\begin{figure*}[h!]
\centering
\includegraphics[width=1.0\linewidth]{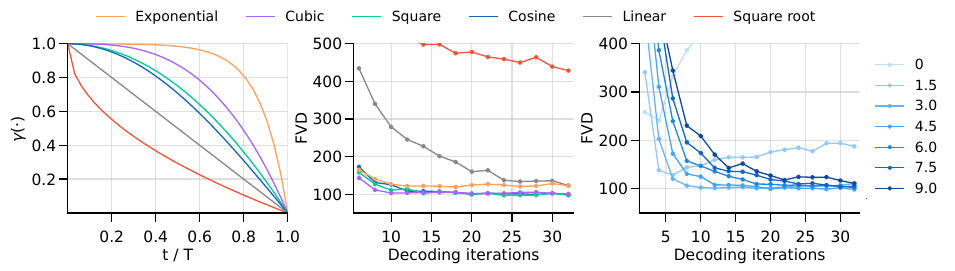}
  \caption{\textbf{Mask scheduling functions.} \textit{Left:} $3$ categories of mask scheduling functions: concave (cosine, square, cubic, exponential), linear, and convex (square root). \textit{Middle:} FVD scores for different mask scheduling functions and decoding iterations. Concave functions perform the best. \textit{Right:} FVD score vs. decoding iterations for different temperature values. Lower and higher temperature values lead to poor FVD scores, with a sweet spot temperature value of $3$ and $4.5$.}
\label{fig:mask_schedule}
\end{figure*}

\subsection{Ablation Studies}
We ablate \mgpt to understand the contribution of each design decision with the default settings: $6$ blocks, $768$ embedding dimension, $1\times16\times16$ spatial window, $16\times4\times4$ spatiotemporal window, and variable masking ratio (Table~\ref{tab:ablations}).

\textbf{Model hyperparameters.} We compare the effect of the number of blocks and the embedding dimension in Table~\ref{tab:model_size}. We find that having a larger embedding dimension improves the performance slightly, whereas increasing the number of blocks does not improve the performance. 

\textbf{Spatiotemporal window (STW).} An important design decision of \mgpt is the size of the STW (Table~\ref{tab:window}). We compare three different window configurations and \mgpt with full self-attention in all layers. Note that training a model with full self-attention requires using gradient checkpointing, which significantly increases the training time. In fact, the STW size of $16\times4\times4$ achieves better accuracy, while requiring $60\%$ less memory and speeds up training time by $3.3\times$.

\textbf{Masking ratio during training.} We find that a fixed masking ratio results in poor video prediction performance (Table~\ref{tab:mask_ratio}). A large masking ratio until a maximum of $95\%$ decreases FVD. Further increase in masking ratio significantly deteriorates the performance. A \textit{variable} masking ratio performs best, as it best approximates the different masking ratios encountered during inference.

\begin{wraptable}{r}{8cm}
\vspace{-0.2cm}
\caption{\textbf{Inference speedup} of \mgpt over auto-regressive generation as measured by the number of forward passes. Iterative decoding in \mgpt can predict video frames in significantly fewer forward passes, especially when conditioned on actions.}
\label{tab:performance}
\centering
\vspace{0.1cm}
\centering
\scriptsize
\begin{tabular}{l c c c c}
 & & \multirow{1}{*}{auto reg.} & \multicolumn{1}{c}{ours} & \\
\multirow{1}{*}{dataset} & \multirow{1}{*}{pred frames} & \# fwd. pass & \# fwd. pass & \multicolumn{1}{c}{speed up}\\
\midrule
BAIR & 15 & 3,840 & 24 & 160$\times$ \\
BAIR w/ act. & 15 & 3,840 & 12 & 320$\times$ \\
KITTI & 25 & 6,400 & 48 & 133$\times$ \\
RoboNet & 10 & 2,560 & 5 & 512$\times$ \\
\end{tabular}
\end{wraptable}

\textbf{Mask scheduling.} The choice of mask scheduling function and the number of decoding iterations during inference has a significant impact on video generation quality (Fig.~\ref{fig:mask_schedule}). We compare three types of scheduling functions: concave (cosine, square, cubic, exponential), linear, and convex (square root). Concave functions performed significantly better than linear and convex functions. Videos have a lot of redundant information due to temporal coherence, and consequently with only $5\%$ of unmasked tokens (Table~\ref{tab:mask_ratio}) the entire video can be correctly predicted. The critical step is to predict these few tokens very accurately. We hypothesize that concave functions perform better as they capture this intuition by slowly predicting the initial tokens over multiple iterations and then rapidly predicting the majority of remaining tokens conditioned on the (more accurate) initial tokens in the final iterations. Convex functions operate in an opposite manner and thus perform significantly worse. Across all functions, FVD improved with increased numbers of decoding steps until a certain point. Further increasing the decoding steps did not improve FVD. Additionally, we found that selecting the most confident tokens while performing iterative decoding led to video predictions with little or no motion. Hence, we add temperature annealed Gumbel noise to the token confidence to encourage the model to produce more diverse outputs (Fig.~\ref{fig:mask_schedule}). Empirically, we found that a temperature value of $4.5$ works best across different datasets. 

\subsection{Visual Model Predictive Control with \mgpt on a Real Robot} 
\label{subsec:robot}

\begin{figure}[t]
\RawFloats
\noindent\begin{minipage}[b]{0.60\textwidth}
    \includegraphics[width=0.49\linewidth]{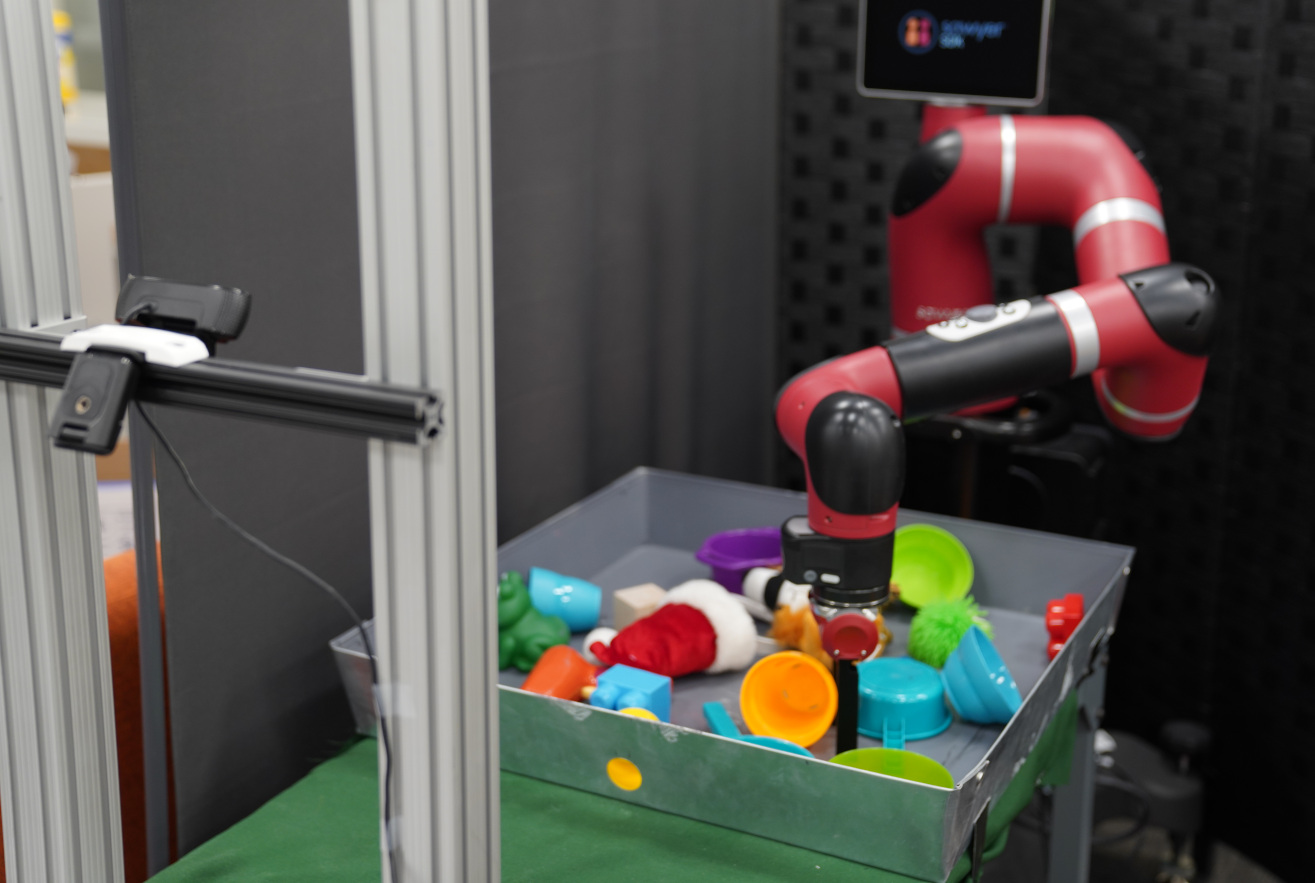}
    \includegraphics[width=0.49\linewidth]{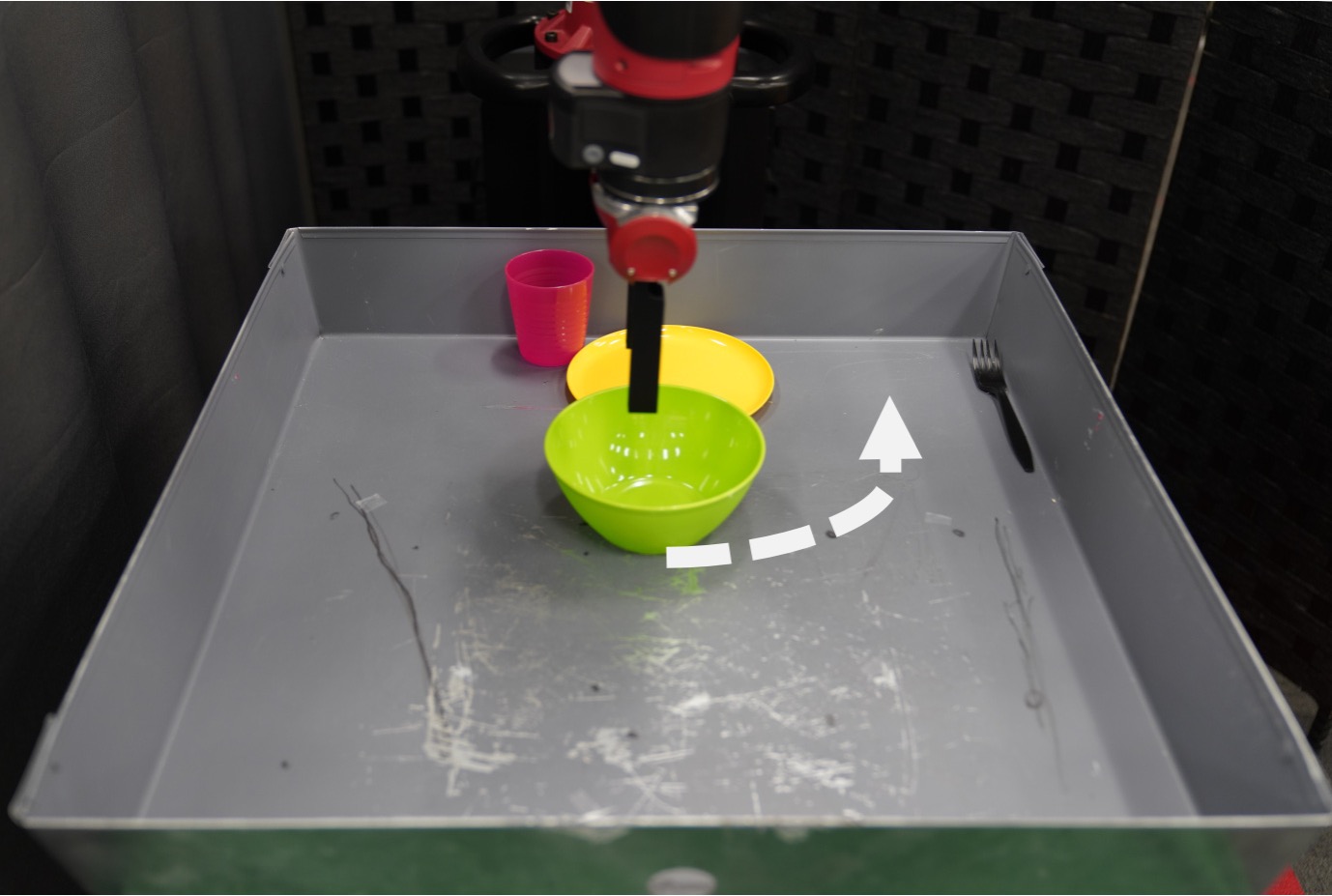}
    \captionof{figure}{\textit{Left}: Third person view of real-world data collection. \textit{Right}: An example evaluation task. The overlaid white arrow depicts the goal location of the green bowl.
    \label{fig:real_robot}}
  \end{minipage}%
  \hfill%
  \begin{minipage}[b]{0.38\textwidth}%
    \centering
    \footnotesize
    \vspace*{\fill}
    \begin{tabular}{lr}
      method & success rate \\ \midrule
        \mgpt (all data) & 60\% \\
        \mgpt (finetuned) & 53\% \\
        \mgpt (RN only) & 3\%\\
        Random policy & 3\% \\
        & \\
        & \\
      \end{tabular}
      \vspace*{\fill}
      \captionof{table}{\textbf{Control evaluation results.} We perform 30 trials for each method, and report aggregated success rates.
      \label{tab:robot_results}}
\end{minipage}%
\end{figure}

Essential to our design is to achieve an inference performance that supports robotics and embodied AI tasks. 
We evaluate whether the performance improvements afforded by our method can enable the control of embodied agents through experimental evaluation on a Sawyer robot arm. 
We first train our model on the RoboNet dataset along with a small collection of random interaction trajectories in our setup. 
We then leverage \mgpt to perform visual model-predictive control, and evaluate the robot’s performance on several manipulation tasks.

\textbf{Setup and data collection.} We autonomously collect $120$K frames of additional finetuning data from our setup to augment RoboNet. Between each pair of frames, the robot takes a $5$-dimensional action representing a change in Cartesian end-effector position: $[x, y, z]$ gripper position, $\theta$ yaw angle, and a binary gripper open/close command.
During data collection, the robot interacts with a diverse collection of random household objects (Fig.~\ref{fig:real_robot}).

\textbf{Model predictive control using \mgpt.} We evaluate the planning capabilities of \mgpt using visual foresight~\cite{finn2017deep, ebert2018visual} on a series of robotic pushing tasks. Our control evaluation contains two task types: \texttt{table setting}, which involves pushing a bowl previously unseen in the training data, and \texttt{sweeping}, where objects are moved into an unseen dustpan. For each task, the robot is given a $64 \times 64$ goal image and we perform planning based on \mgpt by optimizing a sequence of actions using the cross-entropy method (CEM)~\citep{deboer2008cem}, using the $\ell_2$ pixel error between the last predicted image and the goal as the cost.
We evaluate two variants of our model: one trained on the combined RoboNet and finetuning datasets (\texttt{all data}), and another that is pretrained using RoboNet and then finetuned on only the domain-specific data (\texttt{finetuned}). We compare to a baseline model trained only on RoboNet (\texttt{RN only}) as well as a random Gaussian policy. See the supplementary material for additional details and hyperparameters. 

\textbf{Results.} As shown in Table~\ref{tab:robot_results}, our model achieves strong planning performance when provided finetuning data, but the specific method of integrating the finetuning data (all together or post-finetuning) is not significant. We find qualitatively that a model trained only on RoboNet is unable to produce high-fidelity reconstructions of the scene, and cannot predict plausible arm motions. Critically, our method’s efficient inference procedure allows us to achieve $\sim 6.5$ seconds per CEM iteration, which is orders of magnitude more efficient than autoregressive models as shown in Table~\ref{tab:performance}.

%% file: 05-appendix.tex
\appendix

\section{Implementation Details}
\subsection{Training MaskViT}\label{appendix:hyper}
\textbf{VQ-GAN.} We train a VQ-GAN \citep{esser2021taming} model for each dataset which downsamples each frame into $16\times16$ latent codes, i.e., by a factor of $4$ for frames of size $64\times64$ frames and $16$ for frames of size $256\times256$. Table~\ref{tab:hyperparams} summarizes our settings for all three datasets. Training VQ-GAN with discriminator loss can lead to instabilities. Hence, as suggested by \citep{esser2021taming} we start GAN losses after the reconstruction loss has converged. We also found that GAN losses were not always helpful, especially at lower input resolutions for BAIR and RoboNet. 
 
\textbf{Transformer.} Our transformer model is a stack of $L$ blocks, where each block consists of two transformer layers with attention restricted to the window size of $1\times16\times16$ (spatial window) and $T\times4\times4$ (spatiotemporal window), unless otherwise specified. We use learnable positional embeddings, which are the sum of space and time positional embeddings. Following \citep{liu2021swin}, we adopt relative position biases in our layers. We use the Adam \citep{kingma2015adam} optimizer with linear warmup \citep{goyal2017accurate} and a cosine decay learning rate schedule. Table~\ref{tab:hyperparams} summarizes our settings for all three datasets.

\textbf{Evaluation.} We find the optimal evaluation parameters by doing a grid search of the following parameters: $\gamma$ (cosine, square), temperature (3, 4.5) and decoding iterations depending on the prediction horizon length. We use top-p value of $0.95$ for the BAIR dataset only. Table~\ref{tab:hyperparams} summarizes our evaluation settings.

\begin{table*}[h]
    \setlength\tabcolsep{0.5em}
    \centering
    \caption{\textbf{Training and evaluation hyperparameters.}}
    \resizebox{\linewidth}{!}{
        \begin{tabular}{@{}l|ccccc@{}}
            \toprule
            \textbf{Dataset}           & BAIR      & KITTI    & KITTI    & RoboNet     & RoboNet  \\\midrule
            Image resolution           & 64        & 64       & 256      & 64          & 256      \\ 
            Context frames             & 1         & 5        & 5        & 2           & 2        \\  
            \multicolumn{6}{c}{}                                                                  \\\midrule
            \textbf{VQ-GAN}            &           &          &          &             &          \\\midrule
            Channel                    & 160       & 128      & 128      & 192         & 128      \\
            $K$                        & 1024      & 1024     & 1024     & 1024        & 1024     \\
            $n_z$                      & 256       & 256      & 256      & 256         & 256      \\
            Batch size                 & 320       & 1120     & 112      & 720         & 112      \\
            Training steps             & 3e5       & 5e4      & 3e5      & 3e5         & 3e5      \\
            Learning rate              & 1e-4      & 1e-3     & 1e-4     & 5e-4        & 1e-4     \\
            Disc. start                & -         & 2e4      & 1.5e5    & -           & 1.5e5    \\
            \multicolumn{6}{c}{}                                                                  \\\midrule
            \textbf{Transformer}       &           &          &          &             &          \\\midrule
            Spatial window             & \sw       & \sw      & \sw      & \sw         & \sw      \\
            Spatiotemporal window      & \stw{8}   & \stw{4}  & \stw{4}  & \stw{4}     & \stw{4}  \\
            Blocks                     & 6         & 8        & 6        & 8           & 6        \\
            Attention heads            & 4         & 4        & 4        & 4           & 4        \\
            Embedding dim.             & 768       & 768      & 1024     & 768         & 1024     \\
            Feedforward dim.           & 3072      & 3072     & 4096     & 3072        & 4096     \\
            Dropout                    & 0.0       & 0.0      & 0.0      & 0.0         & 0.0      \\
            Batch size                 & 64        & 32       & 32       & 224         & 224      \\
            Learning rate              & 3e-4      & 3e-4     & 3e-4     & 3e-4        & 3e-4     \\
            Training steps             & 1e5       & 1e5      & 1e5      & 3e5         & 3e5      \\
            \multicolumn{6}{c}{}                                                                  \\\midrule
            \textbf{Evaluation}        &           &          &          &             &          \\\midrule
            Mask scheduling func.      & square    & cosine   & cosine   & cosine      & cosine   \\
            Decoding iters.            & 18        & 48       & 64       & 7           & 16       \\
            Temperature                & 4.5       & 3.0      & 4.5      & -           & -        \\
        \end{tabular}
    }
    \label{tab:hyperparams}
\end{table*}

\subsection{Real Robot Experiments}
\label{appendix:real_robot}

\paragraph{Data collection.} Our robot setup consists of a Sawyer robot arm with a Logitech C922 PRO consumer webcam for recording video frames at $640 \times 480$ resolution. All raw image observations are center-cropped to $480 \times 480$ resolution before being resized to $64 \times 64$ for model training and control in our experiments. We autonomously collect $4000$ trajectories of $30$ timesteps. At each step the robot takes a 5-dimensional action representing a change in state of the end-effector: a delta translation in Cartesian space, $[x, y, z]$, for the gripper position in meters, change in $\theta$ yaw angle of the end-effector, and a binary gripper open/close command. Following the action space used to collect the RoboNet dataset, the pitch and roll of the end-effector are kept fixed such that the gripper points with the fingers towards the table surface. Each action within a trajectory is selected independently and each dimension of the action vector is independent of the others, sampled from a diagonal Gaussian distribution, except the gripper open/close command that closes automatically when the $z$-position of the end-effector reaches below a certain threshold to increase the rate of object interaction. The random action distribution is parameterized by $\mathcal{N}(0, \text{diag}([0.035, 0.035, 0.08, \pi/18, 2])$. During data collection, we provide the robot with a diverse set of training objects to interact with. During evaluation, we test on tasks which require the robot to manipulate unseen bowls in one setting and to push training items into an unseen dustpan in another.

\paragraph{Visual-MPC.} Our control strategy is a visual MPC~\citep{finn2017deep, ebert2018visual} procedure. Given a start and goal image $I_0, I_g \in \mathbb{R}^{64\times 64\times 3}$, the objective is to find an optimal sequence of actions to reach the goal observation from the start. The planning objective can be written as: 
$\min_{a_1, a_2, ... a_H} \sum_{i=1}^{H} c_i \| \hat{f}(I_0, a_1, ... a_H)_i - I_g \|^2_2 $
, where $\hat{f}(I_0, a_1, ... a_H)_i$ represents the $i$th predicted frame by the learned video prediction model (\mgpt in our case), and $c_i$ are a sequence of constant hyperparameters that determine the importance of the difference between the predicted frame and the goal for each time step.

We use the cross-entropy method (CEM)~\citep{de2005tutorial} to optimize a sequence of $H = 10$ future actions for this objective. In each planning iteration, we first sample $M=256$ sequences of random actions. We then provide these sequences, together with two consecutive context frames (the previous and current step observations), and one context action (the action taken at the previous step) to \mgpt. Action sequences are sampled according to a multivariate Gaussian distribution. To bias action sampling towards smoother trajectories, the noise samples for actions in a given random trajectory are correlated across time as in \citet{nagabandi2019pddm}. Specifically, given a correlation coefficient hyperparameter $\beta$, we first compute $u^1_i, u^2_i, ... u^M_i \stackrel{i.i.d}{\sim} N(0, \Sigma_i)$, where $\Sigma_i$ is the variance of the action at timestep $i$ in the current optimization iteration. The noise at timestep $i$ for the $j$th random trajectory, $n^j_i$, is then computed as a weighted combination of the new noise sample and the noise sample at the previous timestep, that is, $n^j_i = (1-\beta) * u^j_i + \beta * n^j_{i-1}$. After all noise samples are computed, they are summed with means $\mu_i$ for each timestep, which are also iteratively updated. The final random trajectories are formed by rounding the elements in the last action dimension (gripper action) to $-1$ or $1$, whichever is closer.  

Next, we compare the predictions to the goal image by computing the $\ell_2$ error and summing over time as described by the objective above. We weight the cost on the final timestep by $10\times$, but still include the costs on intermediate timesteps in the summation to encourage the robot to solve the task quickly. The best action sequences based on this score are used to refit the sampling distribution mean and variance for the next optimization iteration.  After $K = 3$ optimization iterations, we execute the best scoring action sequence on the robot for the first $3$ steps before performing replanning. 

The robot uses a total of $15$ steps to solve the task, including one initial action $= [0, 0, -0.08, 0.1, 0]$ which is executed at the beginning of every trajectory. This ensures that at least two context images provided for planning. The planning hyperparameters are summarized in Table~\ref{tab:planning_hyperparameters}.

\textbf{Evaluation.} We perform control evaluation on two categories of tasks: \texttt{table setting} and \texttt{sweeping}. For each task, we test $5$ different variations with $3$ trials each. A trial is considered successful if the center of the object of interest is within $8$ cm of the goal position after the trajectory is complete. Model inference for real robot control is performed using $8$ NVIDIA RTX 3090 GPUs with a batch size of $16$ per GPU. We use $5$ decoding iterations, which yields a forward pass time of approximately $6.2$ seconds for a batch of $256$ samples. 

\begin{table}[h]
    \setlength\tabcolsep{2em}
    \centering
    \begin{tabular}{lr}
        \toprule
        Hyperparameter & Value \\
        \midrule 
        Total trajectory length ($T$)  & $15$ \\
        Planning horizon & $10$ \\
        Number of steps between replanning & $3$ \\
        Action dimension  & $5$ \\
        \# of samples per CEM iteration ($M$) & $256$ \\
        \# of CEM iterations ($K$) & $3$ \\ 
        Weights on each timestep in cost ($c_i$) & $1$ if $i=0, ... 8$; $10$ if $i=9$ \\
        Initial sampling distribution mean & $[0, 0, -0.5, 0, 0]$ \\
        Initial sample distribution std. & $[0.05, 0.05, 0.08, \pi/18, 2]$ \\
        Sampled noise correlation coefficient ($\beta$) & $0.3$ \\
        CEM fraction of elites & $0.05$ \\
        \mgpt mask scheduling function& Cosine \\ 
        \mgpt decoding iterations & $5$ \\ 
    \end{tabular}
    \caption{\textbf{Hyperparameters for visual-MPC.}}
    \label{tab:planning_hyperparameters}
\end{table}

\section{Additional Results}

\subsection{Qualitative Results}\label{appendix:qual_results}
\textbf{Video prediction.} We present additional qualitative video prediction results for BAIR (Fig.~\ref{fig:appendix_bair}), KITTI (Fig.~\ref{fig:appendix_kitti}) and RoboNet (Fig.~\ref{fig:appendix_robonet}).

\textbf{Real robot experiments.} Fig.~\ref{fig:planning_qualitative_task_1} and Fig.~\ref{fig:planning_qualitative_task_2} depict sample predictions for \mgpt (\texttt{all data}) and \mgpt trained only on RoboNet (\texttt{RN only}) for two example control tasks. We observe that with our model, the planner is able to find sequences of actions which bring the blue bowl or the soft red hat close to the position specified in the goal image. However, with a model which is trained only on the RoboNet dataset, planning fails. We see qualitatively that the model trained in RoboNet-only, even when solely performing reconstruction of the first two context images using the VQ-GAN component, is unable to reconstruct the background and robot arm with high fidelity. Despite the diversity of the RoboNet dataset, finetuning on domain-specific data is still required to produce reasonable predictions in our setting.

\subsection{Quantitative Results}
\textbf{Real robot experiments.} Table~\ref{tab:detailed_control_results} shows per-task success rates for our real-world robotic control evaluation. Each of our two task types (table setting, sweeping) has $5$ variations, each of which involves different objects to push (unseen bowls for table setting, toys previously seen in the finetuning data for sweeping) and different target locations. 
\input{realrobot_table}

\begin{landscape}
\centering
\begin{figure}
\includegraphics[width=0.75\textwidth]{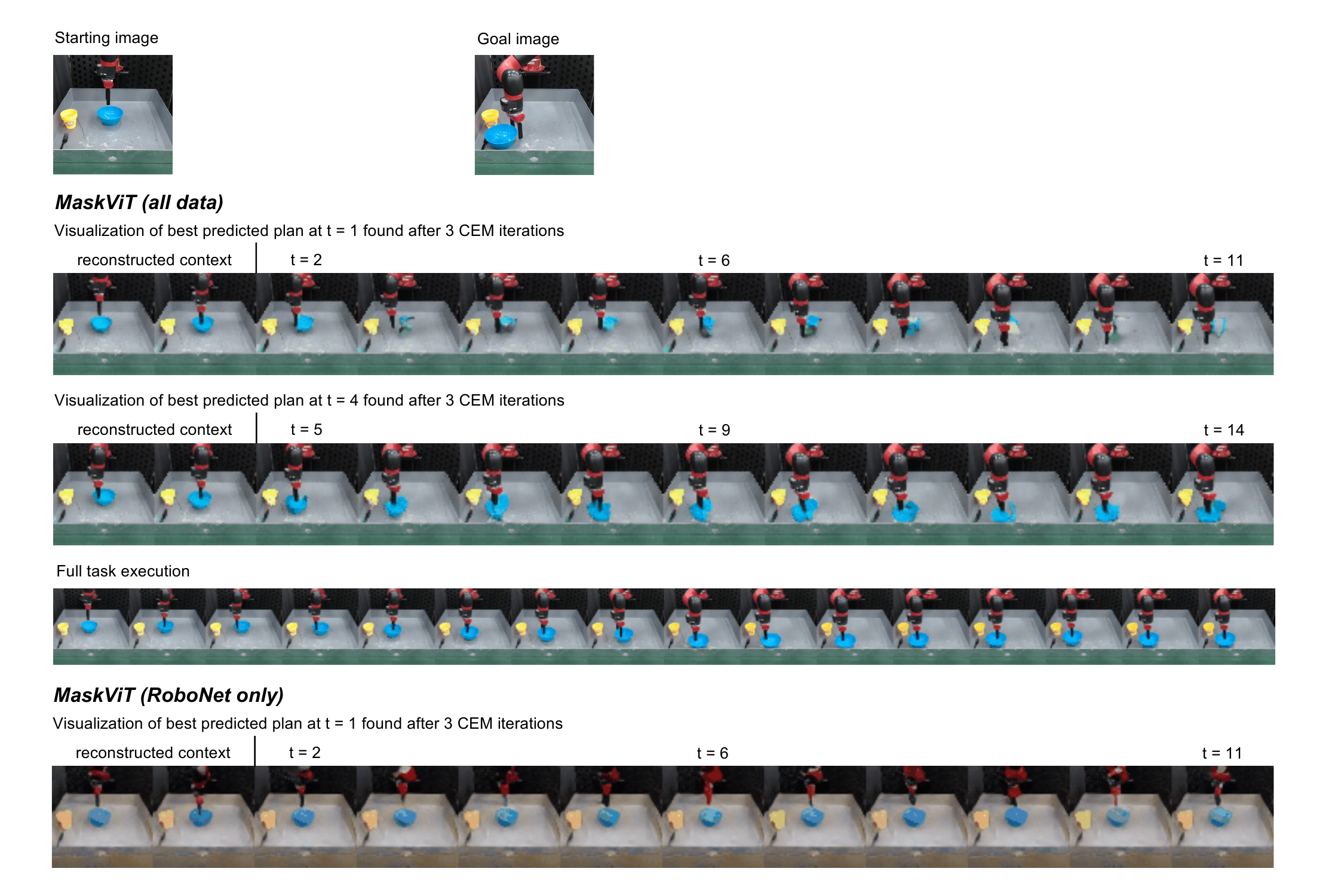}
\caption{Visualizations for task \texttt{set table: push blue bowl left}.
\label{fig:planning_qualitative_task_1}}
\end{figure}

\centering
\begin{figure}
\includegraphics[width=0.75\textwidth]{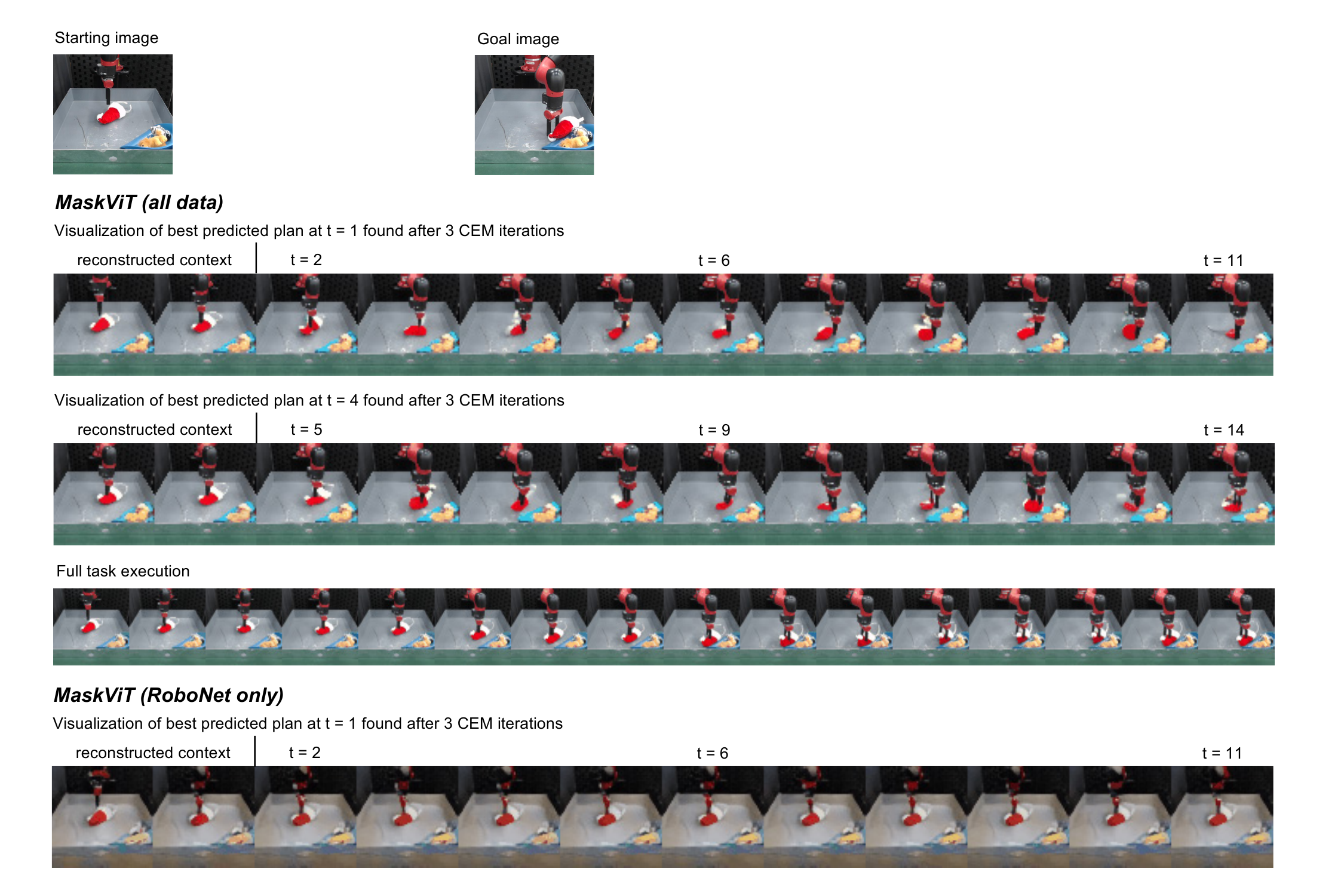}
\caption{Visualizations for task \texttt{sweep: push hat to bottom-right corner}.
\label{fig:planning_qualitative_task_2}}
\end{figure}
\end{landscape}

\begin{landscape}
    \begin{figure}
        \includegraphics[width=0.60\linewidth]{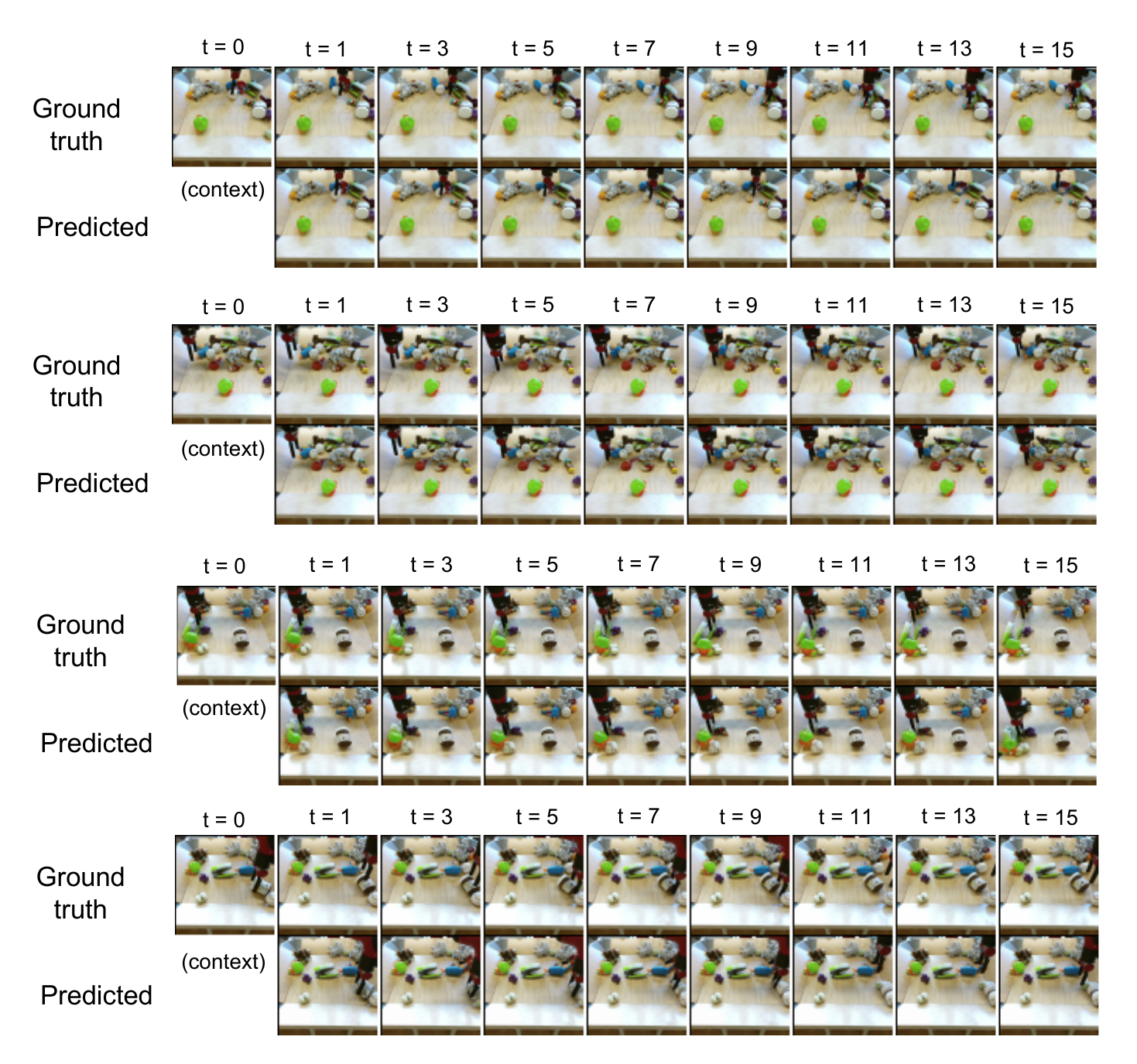}
        \caption{\textbf{Qualitative results: BAIR} (action free, $64\times 64$)}
        \label{fig:appendix_bair}
    \end{figure}
    
    \begin{figure}
        \includegraphics[width=0.65\linewidth]{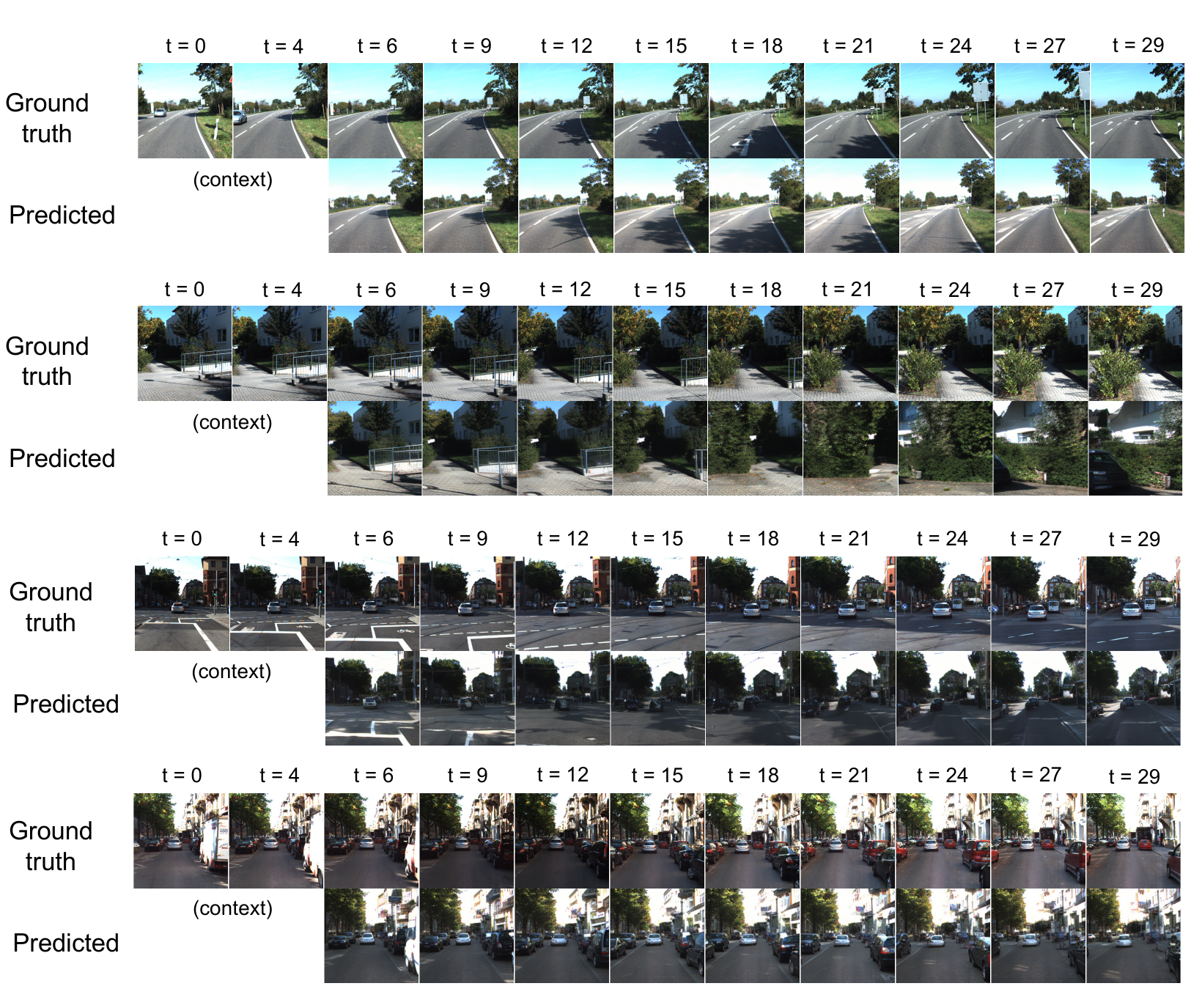}
        \caption{\textbf{Qualitative results: KITTI ($256\times 256$)}}
        \label{fig:appendix_kitti}
    \end{figure}
    
    \begin{figure}
        \includegraphics[width=0.65\linewidth]{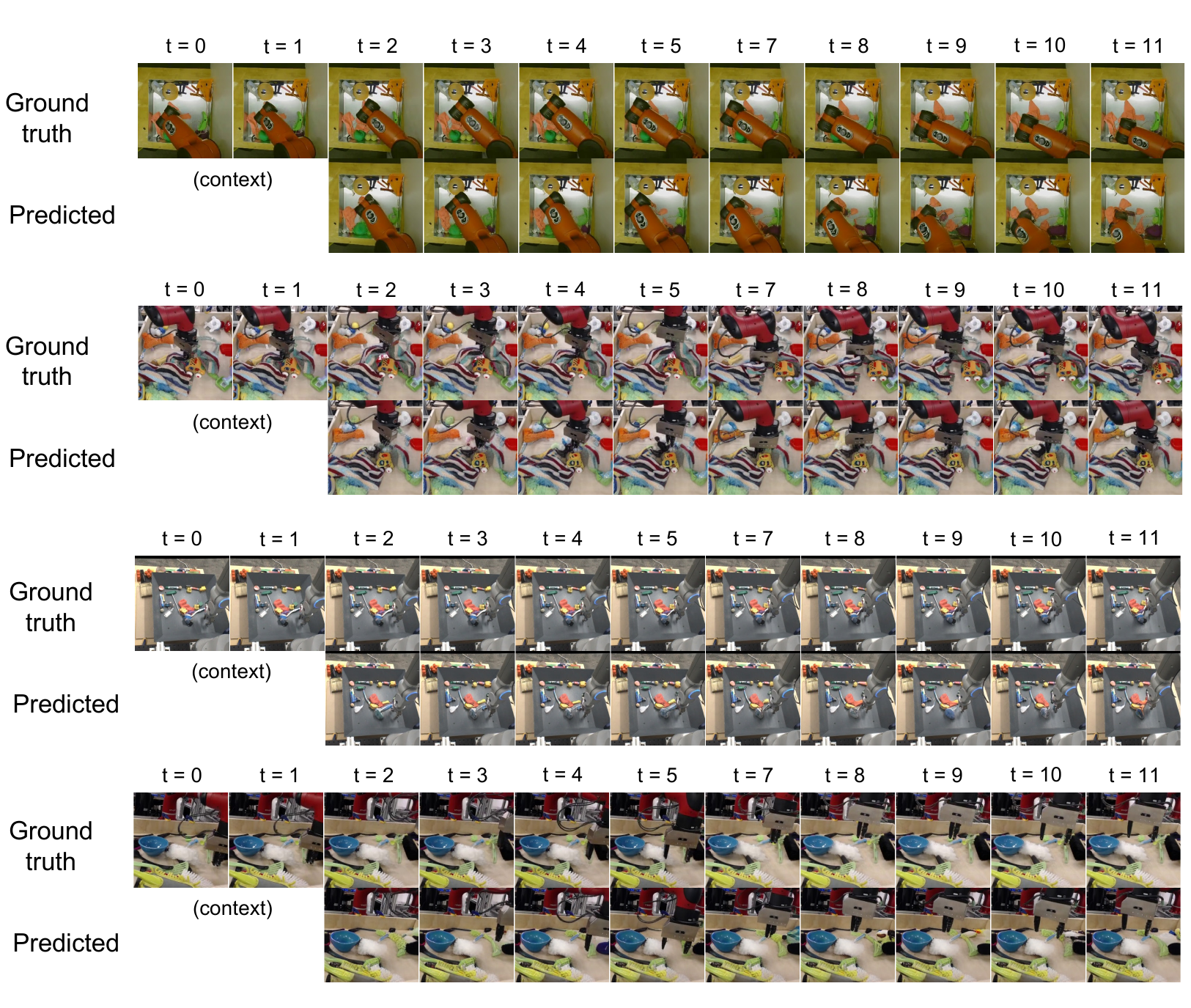}
        \caption{\textbf{Qualitative results: RoboNet ($256 \times 256$)}}
        \label{fig:appendix_robonet}
    \end{figure}
\end{landscape}

%% file: realrobot_table.tex

\newcommand{\ToneVone}{blue; bot-left\xspace}
\newcommand{\ToneVtwo}{green; top-right\xspace}
\newcommand{\ToneVthree}{blue; bot-right \xspace}
\newcommand{\ToneVfour}{red; front\xspace}
\newcommand{\ToneVfive}{green; left\xspace}
\newcommand{\TtwoVone}{toys; top-right\xspace}
\newcommand{\TtwoVtwo}{hat; bot-right\xspace}
\newcommand{\TtwoVthree}{hat; bot-left\xspace}
\newcommand{\TtwoVfour}{toys; top-left\xspace}
\newcommand{\TtwoVfive}{toys; bot-left\xspace}

\definecolor{zero}{rgb}{.75,.75,.75}
\definecolor{one}{rgb}{.5,.5,.5}
\definecolor{two}{rgb}{.25,.25,.25}
\definecolor{two}{rgb}{.0,.0,.0}

\newcommand{\zero}{\cellcolor[rgb]{.45,.45,.45}$0$}
\newcommand{\one}{\cellcolor[rgb]{.6,.6,.6}$1$}
\newcommand{\two}{\cellcolor[rgb]{.85,.85,.85}$2$}
\newcommand{\three}{\cellcolor[rgb]{1.0,1.0,1.0}$3$}

\begin{table*}[h]
    \newcommand*{\tabindent}{\hspace{3mm}}
    \setlength\tabcolsep{0.5em}
    \centering
    \caption{\textbf{Per-task quantitative results} for our robotic control evaluation. We evaluate each of $5$ variants of the two tasks using $3$ trials with each model or policy. Success is determined by the center of the object being within $8$cm of the goal position at the end of $15$ steps.}
    \resizebox{\linewidth}{!}{
        \begin{tabular}{@{}l|cccc@{}}
            \toprule
            \textbf{Task}           & \mgpt (all data) & \mgpt (finetuned) & \mgpt (RN only) & Random \\\midrule
            \textit{table setting (bowl color; destination)} & & & & \\ 
            \tabindent blue; front-left               & $2/3$        & $1/3$        & $0/3$ & $0/3$ \\ 
            \tabindent green; back-right             & $0/3$         &  $0/3$       & $0/3$  & $0/3$  \\  
            \tabindent blue; front-right             & $2/3$         & $2/3$        & $1/3$        & $0/3$  \\  
            \tabindent red; front              & $3/3$         & $2/3$        & $0/3$        & $0/3$  \\  
            \tabindent green; left              & $2/3$        & $3/3$        & $0/3$        & $0/3$  \\  
            \textit{sweeping (object; destination)} & & & & \\ 
            \tabindent toys; back-right               & $3/3$        & $3/3$        & $0/3$ & $0/3$ \\ 
            \tabindent hat; front-right & $1/3$         &  $2/3$       & $0/3$  & $0/3$  \\  
            \tabindent hat; front-left & $2/3$         & $1/3$        & $0/3$        & $1/3$  \\  
            \tabindent toys; back-left & $1/3$         & $1/3$        & $0/3$        & $0/3$  \\  
            \tabindent toys; front-left & $2/3$        & $1/3$        & $0/3$        & $0/3$  \\ \midrule
            Aggregated & $18/30$ & $16/30$ & $1/30$ & $1/30$ \\
        \end{tabular}
    }
    \label{tab:detailed_control_results}
\end{table*}

%% file: main.bbl
\begin{thebibliography}{72}
\providecommand{\natexlab}[1]{#1}
\providecommand{\url}[1]{\texttt{#1}}
\expandafter\ifx\csname urlstyle\endcsname\relax
  \providecommand{\doi}[1]{doi: #1}\else
  \providecommand{\doi}{doi: \begingroup \urlstyle{rm}\Url}\fi

\bibitem[Tanji and Evarts(1976)]{tanji1976anticipatory}
J.~Tanji and E.~V. Evarts.
\newblock Anticipatory activity of motor cortex neurons in relation to
  direction of an intended movement.
\newblock \emph{Journal of neurophysiology}, 39\penalty0 (5):\penalty0
  1062--1068, 1976.

\bibitem[Wolpert et~al.(1995)Wolpert, Ghahramani, and
  Jordan]{wolpert1995internal}
D.~M. Wolpert, Z.~Ghahramani, and M.~I. Jordan.
\newblock An internal model for sensorimotor integration.
\newblock \emph{Science}, 269\penalty0 (5232):\penalty0 1880--1882, 1995.

\bibitem[Wu et~al.(2016)Wu, Dufford, Mackie, Egan, and Fan]{wu2016capacity}
T.~Wu, A.~J. Dufford, M.-A. Mackie, L.~J. Egan, and J.~Fan.
\newblock The capacity of cognitive control estimated from a perceptual
  decision making task.
\newblock \emph{Scientific reports}, 6\penalty0 (1):\penalty0 1--11, 2016.

\bibitem[Finn and Levine(2017)]{finn2017deep}
C.~Finn and S.~Levine.
\newblock Deep visual foresight for planning robot motion.
\newblock In \emph{{IEEE} International Conference on Robotics and Automation},
  2017.

\bibitem[Ebert et~al.(2018)Ebert, Finn, Dasari, Xie, Lee, and
  Levine]{ebert2018visual}
F.~Ebert, C.~Finn, S.~Dasari, A.~Xie, A.~Lee, and S.~Levine.
\newblock Visual foresight: Model-based deep reinforcement learning for
  vision-based robotic control.
\newblock \emph{arXiv preprint arXiv:1812.00568}, 2018.

\bibitem[Hirose et~al.(2019)Hirose, Xia, Mart{\'i}n-Mart{\'i}n, Sadeghian, and
  Savarese]{hirose2019deep}
N.~Hirose, F.~Xia, R.~Mart{\'i}n-Mart{\'i}n, A.~Sadeghian, and S.~Savarese.
\newblock Deep visual mpc-policy learning for navigation.
\newblock \emph{IEEE Robotics and Automation Letters}, 4\penalty0 (4):\penalty0
  3184--3191, 2019.

\bibitem[Vaswani et~al.(2017)Vaswani, Shazeer, Parmar, Uszkoreit, Jones, Gomez,
  Kaiser, and Polosukhin]{vaswani2017attention}
A.~Vaswani, N.~Shazeer, N.~Parmar, J.~Uszkoreit, L.~Jones, A.~N. Gomez,
  {\L}.~Kaiser, and I.~Polosukhin.
\newblock Attention is all you need.
\newblock In \emph{Advances in Neural Information Processing Systems}, 2017.

\bibitem[Devlin et~al.(2019)Devlin, Chang, Lee, and Toutanova]{devlin2019bert}
J.~Devlin, M.-W. Chang, K.~Lee, and K.~Toutanova.
\newblock {BERT}: Pre-training of deep bidirectional transformers for language
  understanding.
\newblock In \emph{Proceedings of the Conference of the North {A}merican
  Chapter of the Association for Computational Linguistics: Human Language
  Technologies}, 2019.

\bibitem[Van Den~Oord and Vinyals(2017)]{van2017neural}
A.~Van Den~Oord and O.~Vinyals.
\newblock Neural discrete representation learning.
\newblock In \emph{Advances in Neural Information Processing Systems}, 2017.

\bibitem[Esser et~al.(2021)Esser, Rombach, and Ommer]{esser2021taming}
P.~Esser, R.~Rombach, and B.~Ommer.
\newblock Taming transformers for high-resolution image synthesis.
\newblock In \emph{Proceedings of the IEEE/CVF Conference on Computer Vision
  and Pattern Recognition}, pages 12873--12883, 2021.

\bibitem[Sohl-Dickstein et~al.(2015)Sohl-Dickstein, Weiss, Maheswaranathan, and
  Ganguli]{sohl2015deep}
J.~Sohl-Dickstein, E.~Weiss, N.~Maheswaranathan, and S.~Ganguli.
\newblock Deep unsupervised learning using nonequilibrium thermodynamics.
\newblock In \emph{International Conference on Machine Learning}, pages
  2256--2265. PMLR, 2015.

\bibitem[Ho et~al.(2020)Ho, Jain, and Abbeel]{ho2020denoising}
J.~Ho, A.~Jain, and P.~Abbeel.
\newblock Denoising diffusion probabilistic models.
\newblock \emph{Advances in Neural Information Processing Systems},
  33:\penalty0 6840--6851, 2020.

\bibitem[Nichol and Dhariwal(2021)]{nichol2021improved}
A.~Q. Nichol and P.~Dhariwal.
\newblock Improved denoising diffusion probabilistic models.
\newblock In \emph{International Conference on Machine Learning}, pages
  8162--8171. PMLR, 2021.

\bibitem[Ghazvininejad et~al.(2019)Ghazvininejad, Levy, Liu, and
  Zettlemoyer]{ghazvininejad2019mask}
M.~Ghazvininejad, O.~Levy, Y.~Liu, and L.~Zettlemoyer.
\newblock Mask-predict: Parallel decoding of conditional masked language
  models.
\newblock In \emph{Proceedings of the 2019 Conference on Empirical Methods in
  Natural Language Processing and the 9th International Joint Conference on
  Natural Language Processing (EMNLP-IJCNLP)}, pages 6112--6121, Hong Kong,
  China, Nov. 2019. Association for Computational Linguistics.
\newblock \doi{10.18653/v1/D19-1633}.

\bibitem[Chang et~al.(2022)Chang, Zhang, Jiang, Liu, and
  Freeman]{chang2022maskgit}
H.~Chang, H.~Zhang, L.~Jiang, C.~Liu, and W.~T. Freeman.
\newblock Maskgit: Masked generative image transformer.
\newblock \emph{arXiv preprint arXiv:2202.04200}, 2022.

\bibitem[Ebert et~al.(2017)Ebert, Finn, Lee, and Levine]{ebert2017self}
F.~Ebert, C.~Finn, A.~X. Lee, and S.~Levine.
\newblock Self-supervised visual planning with temporal skip connections.
\newblock In \emph{Conference on Robot Learning}, pages 344--356, 2017.

\bibitem[Geiger et~al.(2013)Geiger, Lenz, Stiller, and
  Urtasun]{geiger2013vision}
A.~Geiger, P.~Lenz, C.~Stiller, and R.~Urtasun.
\newblock Vision meets robotics: The kitti dataset.
\newblock \emph{The International Journal of Robotics Research}, 32\penalty0
  (11):\penalty0 1231--1237, 2013.

\bibitem[Dasari et~al.(2019)Dasari, Ebert, Tian, Nair, Bucher, Schmeckpeper,
  Singh, Levine, and Finn]{dasari2019robonet}
S.~Dasari, F.~Ebert, S.~Tian, S.~Nair, B.~Bucher, K.~Schmeckpeper, S.~Singh,
  S.~Levine, and C.~Finn.
\newblock Robonet: Large-scale multi-robot learning.
\newblock \emph{arXiv preprint arXiv:1910.11215}, 2019.

\bibitem[Ranzato et~al.(2014)Ranzato, Szlam, Bruna, Mathieu, Collobert, and
  Chopra]{ranzato2014video}
M.~Ranzato, A.~Szlam, J.~Bruna, M.~Mathieu, R.~Collobert, and S.~Chopra.
\newblock Video (language) modeling: a baseline for generative models of
  natural videos.
\newblock \emph{arXiv preprint arXiv:1412.6604}, 2014.

\bibitem[Lotter et~al.(2016)Lotter, Kreiman, and Cox]{lotter2016deep}
W.~Lotter, G.~Kreiman, and D.~Cox.
\newblock Deep predictive coding networks for video prediction and unsupervised
  learning.
\newblock \emph{arXiv preprint arXiv:1605.08104}, 2016.

\bibitem[Li et~al.(2018)Li, Min, Shen, Carlson, and Carin]{li2018video}
Y.~Li, M.~Min, D.~Shen, D.~Carlson, and L.~Carin.
\newblock Video generation from text.
\newblock In \emph{Proceedings of the AAAI Conference on Artificial
  Intelligence}, 2018.

\bibitem[Gupta et~al.(2018)Gupta, Schwenk, Farhadi, Hoiem, and
  Kembhavi]{gupta2018imagine}
T.~Gupta, D.~Schwenk, A.~Farhadi, D.~Hoiem, and A.~Kembhavi.
\newblock Imagine this! scripts to compositions to videos.
\newblock In \emph{Proceedings of the European Conference on Computer Vision
  (ECCV)}, pages 598--613, 2018.

\bibitem[Pan et~al.(2017)Pan, Qiu, Yao, Li, and Mei]{pan2017create}
Y.~Pan, Z.~Qiu, T.~Yao, H.~Li, and T.~Mei.
\newblock To create what you tell: Generating videos from captions.
\newblock In \emph{Proceedings of the 25th ACM International Conference on
  Multimedia}, pages 1789--1798, 2017.

\bibitem[Wu et~al.(2021)Wu, Liang, Ji, Yang, Fang, Jiang, and Duan]{wu2021n}
C.~Wu, J.~Liang, L.~Ji, F.~Yang, Y.~Fang, D.~Jiang, and N.~Duan.
\newblock {NUWA}: Visual synthesis pre-training for neural visual world
  creation.
\newblock \emph{arXiv preprint arXiv:2111.12417}, 2021.

\bibitem[Finn et~al.(2016)Finn, Goodfellow, and Levine]{finn2016unsupervised}
C.~Finn, I.~Goodfellow, and S.~Levine.
\newblock Unsupervised learning for physical interaction through video
  prediction.
\newblock In \emph{Advances in Neural Information Processing Systems}, 2016.

\bibitem[Villegas et~al.(2019)Villegas, Pathak, Kannan, Erhan, Le, and
  Lee]{villegas2019high}
R.~Villegas, A.~Pathak, H.~Kannan, D.~Erhan, Q.~V. Le, and H.~Lee.
\newblock High fidelity video prediction with large stochastic recurrent neural
  networks.
\newblock \emph{Advances in Neural Information Processing Systems}, 32, 2019.

\bibitem[Wu et~al.(2021)Wu, Nair, Martin-Martin, Fei-Fei, and
  Finn]{wu2021greedy}
B.~Wu, S.~Nair, R.~Martin-Martin, L.~Fei-Fei, and C.~Finn.
\newblock Greedy hierarchical variational autoencoders for large-scale video
  prediction.
\newblock In \emph{Proceedings of the IEEE/CVF Conference on Computer Vision
  and Pattern Recognition}, pages 2318--2328, 2021.

\bibitem[Babaeizadeh et~al.(2021)Babaeizadeh, Saffar, Nair, Levine, Finn, and
  Erhan]{babaeizadeh2021fitvid}
M.~Babaeizadeh, M.~T. Saffar, S.~Nair, S.~Levine, C.~Finn, and D.~Erhan.
\newblock Fitvid: Overfitting in pixel-level video prediction.
\newblock \emph{arXiv preprint arXiv:2106.13195}, 2021.

\bibitem[Clark et~al.(2019)Clark, Donahue, and Simonyan]{clark2019adversarial}
A.~Clark, J.~Donahue, and K.~Simonyan.
\newblock Adversarial video generation on complex datasets.
\newblock \emph{arXiv preprint arXiv:1907.06571}, 2019.

\bibitem[Tulyakov et~al.(2018)Tulyakov, Liu, Yang, and
  Kautz]{tulyakov2018mocogan}
S.~Tulyakov, M.-Y. Liu, X.~Yang, and J.~Kautz.
\newblock Mocogan: Decomposing motion and content for video generation.
\newblock In \emph{Proceedings of the IEEE Conference on Computer Vision and
  Pattern Recognition}, pages 1526--1535, 2018.

\bibitem[Luc et~al.(2020)Luc, Clark, Dieleman, Casas, Doron, Cassirer, and
  Simonyan]{luc2020transformation}
P.~Luc, A.~Clark, S.~Dieleman, D.~d.~L. Casas, Y.~Doron, A.~Cassirer, and
  K.~Simonyan.
\newblock Transformation-based adversarial video prediction on large-scale
  data.
\newblock \emph{arXiv preprint arXiv:2003.04035}, 2020.

\bibitem[Babaeizadeh et~al.(2018)Babaeizadeh, Finn, Erhan, Campbell, and
  Levine]{babaeizadeh2018stochastic}
M.~Babaeizadeh, C.~Finn, D.~Erhan, R.~H. Campbell, and S.~Levine.
\newblock Stochastic variational video prediction.
\newblock In \emph{International Conference on Learning Representations}, 2018.

\bibitem[Denton and Fergus(2018)]{denton2018stochastic}
E.~Denton and R.~Fergus.
\newblock Stochastic video generation with a learned prior.
\newblock In \emph{International Conference on Machine Learning}, pages
  1174--1183. PMLR, 2018.

\bibitem[Akan et~al.(2021)Akan, Erdem, Erdem, and G{\"u}ney]{akan2021slamp}
A.~K. Akan, E.~Erdem, A.~Erdem, and F.~G{\"u}ney.
\newblock Slamp: Stochastic latent appearance and motion prediction.
\newblock In \emph{Proceedings of the IEEE/CVF International Conference on
  Computer Vision}, pages 14728--14737, 2021.

\bibitem[Akan et~al.(2022)Akan, Safadoust, Erdem, Erdem, and
  G{\"u}ney]{akan2022stochastic}
A.~K. Akan, S.~Safadoust, E.~Erdem, A.~Erdem, and F.~G{\"u}ney.
\newblock Stochastic video prediction with structure and motion.
\newblock \emph{arXiv preprint arXiv:2203.10528}, 2022.

\bibitem[Dorkenwald et~al.(2021)Dorkenwald, Milbich, Blattmann, Rombach,
  Derpanis, and Ommer]{dorkenwald2021stochastic}
M.~Dorkenwald, T.~Milbich, A.~Blattmann, R.~Rombach, K.~G. Derpanis, and
  B.~Ommer.
\newblock Stochastic image-to-video synthesis using cinns.
\newblock In \emph{Proceedings of the IEEE/CVF Conference on Computer Vision
  and Pattern Recognition}, pages 3742--3753, 2021.

\bibitem[Yan et~al.(2021)Yan, Zhang, Abbeel, and Srinivas]{yan2021videogpt}
W.~Yan, Y.~Zhang, P.~Abbeel, and A.~Srinivas.
\newblock Videogpt: Video generation using vq-vae and transformers.
\newblock \emph{arXiv preprint arXiv:2104.10157}, 2021.

\bibitem[Rakhimov et~al.(2020)Rakhimov, Volkhonskiy, Artemov, Zorin, and
  Burnaev]{rakhimov2020latent}
R.~Rakhimov, D.~Volkhonskiy, A.~Artemov, D.~Zorin, and E.~Burnaev.
\newblock Latent video transformer.
\newblock \emph{arXiv preprint arXiv:2006.10704}, 2020.

\bibitem[Nash et~al.(2022)Nash, Carreira, Walker, Barr, Jaegle, Malinowski, and
  Battaglia]{nash2022transframer}
C.~Nash, J.~Carreira, J.~Walker, I.~Barr, A.~Jaegle, M.~Malinowski, and
  P.~Battaglia.
\newblock Transframer: Arbitrary frame prediction with generative models.
\newblock \emph{arXiv preprint arXiv:2203.09494}, 2022.

\bibitem[Ho et~al.(2022)Ho, Salimans, Gritsenko, Chan, Norouzi, and
  Fleet]{ho2022video}
J.~Ho, T.~Salimans, A.~Gritsenko, W.~Chan, M.~Norouzi, and D.~J. Fleet.
\newblock Video diffusion models.
\newblock \emph{arXiv preprint arXiv:2204.03458}, 2022.

\bibitem[Voleti et~al.(2022)Voleti, Jolicoeur-Martineau, and
  Pal]{voleti2022MCVD}
V.~Voleti, A.~Jolicoeur-Martineau, and C.~Pal.
\newblock Masked conditional video diffusion for prediction, generation, and
  interpolation.
\newblock \emph{arXiv preprint arXiv:2205.09853}, 2022.

\bibitem[Vincent et~al.(2008)Vincent, Larochelle, Bengio, and
  Manzagol]{vincent2008extracting}
P.~Vincent, H.~Larochelle, Y.~Bengio, and P.-A. Manzagol.
\newblock Extracting and composing robust features with denoising autoencoders.
\newblock In \emph{Proceedings of the 25th International Conference on Machine
  Learning}, pages 1096--1103, 2008.

\bibitem[Brown et~al.(2020)Brown, Mann, Ryder, Subbiah, Kaplan, Dhariwal,
  Neelakantan, Shyam, Sastry, Askell, et~al.]{brown2020language}
T.~Brown, B.~Mann, N.~Ryder, M.~Subbiah, J.~D. Kaplan, P.~Dhariwal,
  A.~Neelakantan, P.~Shyam, G.~Sastry, A.~Askell, et~al.
\newblock Language models are few-shot learners.
\newblock In \emph{Advances in Neural Information Processing Systems}, 2020.

\bibitem[Radford et~al.(2019)Radford, Wu, Child, Luan, Amodei, Sutskever,
  et~al.]{radford2019language}
A.~Radford, J.~Wu, R.~Child, D.~Luan, D.~Amodei, I.~Sutskever, et~al.
\newblock Language models are unsupervised multitask learners.
\newblock \emph{OpenAI blog}, 1\penalty0 (8):\penalty0 9, 2019.

\bibitem[He et~al.(2021)He, Chen, Xie, Li, Doll{\'a}r, and
  Girshick]{he2021masked}
K.~He, X.~Chen, S.~Xie, Y.~Li, P.~Doll{\'a}r, and R.~Girshick.
\newblock Masked autoencoders are scalable vision learners.
\newblock \emph{arXiv preprint arXiv:2111.06377}, 2021.

\bibitem[Dosovitskiy et~al.(2020)Dosovitskiy, Beyer, Kolesnikov, Weissenborn,
  Zhai, Unterthiner, Dehghani, Minderer, Heigold, Gelly,
  et~al.]{dosovitskiy2020image}
A.~Dosovitskiy, L.~Beyer, A.~Kolesnikov, D.~Weissenborn, X.~Zhai,
  T.~Unterthiner, M.~Dehghani, M.~Minderer, G.~Heigold, S.~Gelly, et~al.
\newblock An image is worth 16x16 words: Transformers for image recognition at
  scale.
\newblock \emph{arXiv preprint arXiv:2010.11929}, 2020.

\bibitem[Bao et~al.(2022)Bao, Dong, Piao, and Wei]{bao2022beit}
H.~Bao, L.~Dong, S.~Piao, and F.~Wei.
\newblock {BE}it: {BERT} pre-training of image transformers.
\newblock In \emph{International Conference on Learning Representations}, 2022.

\bibitem[Chen et~al.(2020)Chen, Radford, Child, Wu, Jun, Luan, and
  Sutskever]{chen2020generative}
M.~Chen, A.~Radford, R.~Child, J.~Wu, H.~Jun, D.~Luan, and I.~Sutskever.
\newblock Generative pretraining from pixels.
\newblock In \emph{International Conference on Machine Learning}, pages
  1691--1703. PMLR, 2020.

\bibitem[Tong et~al.(2022)Tong, Song, Wang, and Wang]{tong2022videomae}
Z.~Tong, Y.~Song, J.~Wang, and L.~Wang.
\newblock Videomae: Masked autoencoders are data-efficient learners for
  self-supervised video pre-training.
\newblock \emph{arXiv preprint arXiv:2203.12602}, 2022.

\bibitem[Feichtenhofer et~al.(2022)Feichtenhofer, Fan, Li, and
  He]{feichtenhofer2022masked}
C.~Feichtenhofer, H.~Fan, Y.~Li, and K.~He.
\newblock Masked autoencoders as spatiotemporal learners.
\newblock \emph{arXiv preprint arXiv:2205.09113}, 2022.

\bibitem[Laskin et~al.(2020)Laskin, Srinivas, and Abbeel]{laskin2020curl}
M.~Laskin, A.~Srinivas, and P.~Abbeel.
\newblock {CURL}: Contrastive unsupervised representations for reinforcement
  learning.
\newblock In H.~D. III and A.~Singh, editors, \emph{Proceedings of the 37th
  International Conference on Machine Learning}, volume 119 of
  \emph{Proceedings of Machine Learning Research}, pages 5639--5650. PMLR,
  13--18 Jul 2020.

\bibitem[Nair et~al.(2022)Nair, Rajeswaran, Kumar, Finn, and
  Gupta]{nair2022r3m}
S.~Nair, A.~Rajeswaran, V.~Kumar, C.~Finn, and A.~Gupta.
\newblock R3m: A universal visual representation for robot manipulation.
\newblock \emph{arXiv preprint arXiv:2203.12601}, 2022.

\bibitem[Parisi et~al.(2022)Parisi, Rajeswaran, Purushwalkam, and
  Gupta]{parisi2022unsurprising}
S.~Parisi, A.~Rajeswaran, S.~Purushwalkam, and A.~Gupta.
\newblock The unsurprising effectiveness of pre-trained vision models for
  control.
\newblock \emph{arXiv preprint arXiv:2203.03580}, 2022.

\bibitem[Xiao et~al.(2022)Xiao, Radosavovic, Darrell, and
  Malik]{xiao2022masked}
T.~Xiao, I.~Radosavovic, T.~Darrell, and J.~Malik.
\newblock Masked visual pre-training for motor control.
\newblock \emph{arXiv preprint arXiv:2203.06173}, 2022.

\bibitem[Razavi et~al.(2019)Razavi, Van~den Oord, and
  Vinyals]{razavi2019generating}
A.~Razavi, A.~Van~den Oord, and O.~Vinyals.
\newblock Generating diverse high-fidelity images with vq-vae-2.
\newblock In \emph{Advances in Neural Information Processing Systems}, 2019.

\bibitem[Goodfellow et~al.(2014)Goodfellow, Pouget-Abadie, Mirza, Xu,
  Warde-Farley, Ozair, Courville, and Bengio]{goodfellow2014generative}
I.~Goodfellow, J.~Pouget-Abadie, M.~Mirza, B.~Xu, D.~Warde-Farley, S.~Ozair,
  A.~Courville, and Y.~Bengio.
\newblock Generative adversarial nets.
\newblock In \emph{Advances in Neural Information Processing Systems}, 2014.

\bibitem[Johnson et~al.(2016)Johnson, Alahi, and
  Fei-Fei]{johnson2016perceptual}
J.~Johnson, A.~Alahi, and L.~Fei-Fei.
\newblock Perceptual losses for real-time style transfer and super-resolution.
\newblock In \emph{European Conference on Computer Vision}, pages 694--711.
  Springer, 2016.

\bibitem[Zhang et~al.(2018)Zhang, Isola, Efros, Shechtman, and
  Wang]{zhang2018unreasonable}
R.~Zhang, P.~Isola, A.~A. Efros, E.~Shechtman, and O.~Wang.
\newblock The unreasonable effectiveness of deep features as a perceptual
  metric.
\newblock In \emph{Proceedings of the IEEE Conference on Computer Vision and
  Pattern Recognition}, pages 586--595, 2018.

\bibitem[Liu et~al.(2021{\natexlab{a}})Liu, Lin, Cao, Hu, Wei, Zhang, Lin, and
  Guo]{liu2021swin}
Z.~Liu, Y.~Lin, Y.~Cao, H.~Hu, Y.~Wei, Z.~Zhang, S.~Lin, and B.~Guo.
\newblock Swin transformer: Hierarchical vision transformer using shifted
  windows.
\newblock In \emph{Proceedings of the IEEE/CVF International Conference on
  Computer Vision}, pages 10012--10022, 2021{\natexlab{a}}.

\bibitem[Liu et~al.(2021{\natexlab{b}})Liu, Ning, Cao, Wei, Zhang, Lin, and
  Hu]{liu2021video}
Z.~Liu, J.~Ning, Y.~Cao, Y.~Wei, Z.~Zhang, S.~Lin, and H.~Hu.
\newblock Video swin transformer.
\newblock \emph{arXiv preprint arXiv:2106.13230}, 2021{\natexlab{b}}.

\bibitem[Child et~al.(2019)Child, Gray, Radford, and
  Sutskever]{child2019generating}
R.~Child, S.~Gray, A.~Radford, and I.~Sutskever.
\newblock Generating long sequences with sparse transformers.
\newblock \emph{arXiv preprint arXiv:1904.10509}, 2019.

\bibitem[Lee et~al.(2018)Lee, Zhang, Ebert, Abbeel, Finn, and
  Levine]{lee2018stochastic}
A.~X. Lee, R.~Zhang, F.~Ebert, P.~Abbeel, C.~Finn, and S.~Levine.
\newblock Stochastic adversarial video prediction.
\newblock \emph{arXiv preprint arXiv:1804.01523}, 2018.

\bibitem[Weissenborn et~al.(2020)Weissenborn, Täckström, and
  Uszkoreit]{Weissenborn2020Scaling}
D.~Weissenborn, O.~Täckström, and J.~Uszkoreit.
\newblock Scaling autoregressive video models.
\newblock In \emph{International Conference on Learning Representations}, 2020.

\bibitem[Unterthiner et~al.(2018)Unterthiner, van Steenkiste, Kurach, Marinier,
  Michalski, and Gelly]{unterthiner2018towards}
T.~Unterthiner, S.~van Steenkiste, K.~Kurach, R.~Marinier, M.~Michalski, and
  S.~Gelly.
\newblock Towards accurate generative models of video: A new metric \&
  challenges.
\newblock \emph{arXiv preprint arXiv:1812.01717}, 2018.

\bibitem[Wang et~al.(2004)Wang, Bovik, Sheikh, and Simoncelli]{wang2004image}
Z.~Wang, A.~C. Bovik, H.~R. Sheikh, and E.~P. Simoncelli.
\newblock Image quality assessment: from error visibility to structural
  similarity.
\newblock \emph{IEEE Transactions on Image Processing}, 13\penalty0
  (4):\penalty0 600--612, 2004.

\bibitem[Boer et~al.(2004)Boer, Kroese, Mannor, and Rubinstein]{deboer2008cem}
P.~D. Boer, Kroese, S.~Mannor, and R.~Y. Rubinstein.
\newblock A tutorial on the cross-entropy method.
\newblock \emph{Annals of Operations Research}, 134\penalty0 (1):\penalty0
  19--67, 2004.

\bibitem[Grauman et~al.(2021)Grauman, Westbury, Byrne, Chavis, Furnari,
  Girdhar, Hamburger, Jiang, Liu, Liu, et~al.]{grauman2021ego4d}
K.~Grauman, A.~Westbury, E.~Byrne, Z.~Chavis, A.~Furnari, R.~Girdhar,
  J.~Hamburger, H.~Jiang, M.~Liu, X.~Liu, et~al.
\newblock Ego4d: Around the world in 3,000 hours of egocentric video.
\newblock \emph{arXiv preprint arXiv:2110.07058}, 3, 2021.

\bibitem[Srivastava et~al.(2022)Srivastava, Li, Lingelbach,
  Mart{\'i}n-Mart{\'i}n, Xia, Vainio, Lian, Gokmen, Buch, Liu,
  et~al.]{srivastava2022behavior}
S.~Srivastava, C.~Li, M.~Lingelbach, R.~Mart{\'i}n-Mart{\'i}n, F.~Xia, K.~E.
  Vainio, Z.~Lian, C.~Gokmen, S.~Buch, K.~Liu, et~al.
\newblock Behavior: Benchmark for everyday household activities in virtual,
  interactive, and ecological environments.
\newblock In \emph{Conference on Robot Learning}, pages 477--490. PMLR, 2022.

\bibitem[Kingma and Ba(2015)]{kingma2015adam}
D.~P. Kingma and J.~Ba.
\newblock Adam: A method for stochastic optimization.
\newblock In \emph{ICLR (Poster)}, 2015.
\newblock URL \url{http://arxiv.org/abs/1412.6980}.

\bibitem[Goyal et~al.(2017)Goyal, Doll{\'a}r, Girshick, Noordhuis, Wesolowski,
  Kyrola, Tulloch, Jia, and He]{goyal2017accurate}
P.~Goyal, P.~Doll{\'a}r, R.~Girshick, P.~Noordhuis, L.~Wesolowski, A.~Kyrola,
  A.~Tulloch, Y.~Jia, and K.~He.
\newblock Accurate, large minibatch sgd: Training imagenet in 1 hour.
\newblock \emph{arXiv preprint arXiv:1706.02677}, 2017.

\bibitem[De~Boer et~al.(2005)De~Boer, Kroese, Mannor, and
  Rubinstein]{de2005tutorial}
P.-T. De~Boer, D.~P. Kroese, S.~Mannor, and R.~Y. Rubinstein.
\newblock A tutorial on the cross-entropy method.
\newblock \emph{Annals of operations research}, 134\penalty0 (1):\penalty0
  19--67, 2005.

\bibitem[Nagabandi et~al.(2019)Nagabandi, Konoglie, Levine, and
  Kumar]{nagabandi2019pddm}
A.~Nagabandi, K.~Konoglie, S.~Levine, and V.~Kumar.
\newblock {Deep Dynamics Models for Learning Dexterous Manipulation}.
\newblock In \emph{Conference on Robot Learning (CoRL)}, 2019.

\end{thebibliography}
